%% file: Main.tex
\documentclass[letterpaper]{article} 
\usepackage[]{aaai23}  
\usepackage{times}  
\usepackage{helvet}  
\usepackage{courier}  
\usepackage[hyphens]{url}  
\usepackage{graphicx} 
\usepackage{natbib}  
\usepackage{caption} 
\usepackage{booktabs} 
\usepackage{enumerate}
\usepackage{amsmath}
\usepackage{amsfonts}
\usepackage{graphicx}
\usepackage[ruled,vlined]{algorithm2e}
\usepackage{algpseudocode}
\usepackage{bm}
\usepackage{silence}
\usepackage{subfigure}
\usepackage{multirow}
\usepackage{enumitem}
\usepackage{array}
\usepackage{xcolor}
\usepackage{balance}
\usepackage{natbib}

\makeatletter
\renewcommand{\ALG@beginalgorithmic}{\small}
\makeatother

\newtheorem{problem}{Problem Definition}

\title{AAAI Press Anonymous Submission\\Instructions for Authors Using \LaTeX{}}
\author{
    Kaize Ding\textsuperscript{\rm 1},
    Jianling Wang\textsuperscript{\rm 2},
    Jundong Li\textsuperscript{\rm 3},
    James Caverlee\textsuperscript{\rm 2},
    \textrm{and} Huan Liu\textsuperscript{\rm 1}
}
\affiliations{
    \textsuperscript{\rm 1} Arizona State University\\
    \textsuperscript{\rm 2} Texas A\&M University\\
    \textsuperscript{\rm 3} University of Virginia\\
    kaize.ding@asu.edu, jlwang@tamu.edu, jundong@virginia.edu, caverlee@tamu.edu, huan.liu@asu.edu
}

\usepackage{bibentry}

\begin{document}
\title{Robust Graph Meta-learning for Weakly-supervised Few-shot Node Classification}







\maketitle

\begin{abstract}
By transferring the knowledge learned from previous experiences to new tasks, graph meta-learning approaches have demonstrated promising performance on few-shot graph learning problems. However, most existing efforts predominately assume that all the data from the seen classes is gold-labeled, while those methods may lose their efficacy when the seen data is weakly-labeled with severe label noise. As such, we aim to investigate a novel problem of weakly-supervised graph meta-learning for improving the model robustness in terms of knowledge transfer. To achieve this goal, we propose a new graph meta-learning framework -- Graph Interpolation Networks (Meta-GIN). Based on a new robustness-enhanced episodic training paradigm, Meta-GIN is meta-learned to interpolate node representations from weakly-labeled data and extracts highly transferable meta-knowledge, which enables the model to quickly adapt to unseen tasks with few labeled instances. Extensive experiments demonstrate the superiority of Meta-GIN over existing graph meta-learning studies on the task of weakly-supervised few-shot node classification.
\end{abstract}

\section{Introduction}

\input{Intro}

\section{Related Work}

\input{Related.tex}


\section{Proposed Approach}
\input{Methodology.tex}

\section{Experiments}

\input{Experiments.tex}

\section{Conclusion}
\input{Conclusion.tex}



\balance
\bibliography{acmart}

\newpage
\balance
\appendix
\input{Appendix.tex}

\end{document}

%% file: Intro.tex
Many prevailing graph ML methods typically rely upon the availability of sufficient labeled data~\cite{zhou2019meta, ding2020graph}. In contrast, the long-tail property of real-world graphs makes those methods less effective for learning new concepts when only limited labeled data is available~\cite{ding2020graph,huang2020graph,baek2020learning}. A powerful graph ML model should be able to quickly learn never-before-seen class labels using only a handful of labeled data. Dealing with such \textit{few-shot} concepts\footnote{In this paper, we primarily focus on the task of few-shot node classification.} is important and corresponds to many practical applications. For example, many social networking platforms such as Facebook and Twitter need to consistently promote new features or new social media groups to users. Based on the limited user interactions, the deployed model is required to provide high-quality recommendations for other users regarding these new concepts. 
Inspired by the recent success of meta-learning in image domain~\cite{snell2017prototypical}, increasing research efforts have been made in graph meta-learning~\cite{zhou2019meta,huang2020graph,lan2020node,liu2021relative} for solving the problem of few-shot learning on graph-structured data.


Despite some exciting progress, the research of graph meta-learning overall remains in its infancy. In general, existing endeavors predominately focus on the \textit{supervised} setting, where abundant gold-labeled nodes can be accessed from the seen classes during the meta-training process. This assumption is often infeasible since collecting such auxiliary knowledge is laborious and requires intensive domain-knowledge, especially when considering the heterogeneity of graph-structured data. 
An alternative solution is to adopt automatic labeling tools based on heuristics, crowd-sourcing, or weak-learners~\cite{zhang2019metacleaner}. Though using such \textit{weakly-labeled} data is more practical, one companion issue is that those labels usually contain a significant amount of \textit{noise}. As shown in the previous research studies~\cite{ren2018meta,zhang2019variational}, most of the existing few-shot learning (FSL) models are highly vulnerable to noise or outliers, thus the model performance on unseen tasks would be largely degraded if the model is meta-learned on those weakly-labeled data. Therefore, it is imperative to investigate the problem of \textit{weakly-supervised} graph meta-learning, in order to push forward the frontiers of FSL on graphs.

As a research problem has been little explored, weakly-supervised graph meta-learning is challenging to solve, mainly due to the difficulty of \textit{extracting highly transferable meta-knowledge from weakly-labeled node classes}: \textbf{(i)} on the one hand, the existing literature of learning with noisy labels is tailored for independent and identically distributed (\textit{i.i.d.}) data such as image and text. The inability of modeling relational data like graphs that lie in non-Euclidean space could largely jeopardize their effectiveness. Hence, it is necessary to develop new frameworks that can consider the dependencies among nodes and mitigate the inaccurate supervision signals; \textbf{(ii)} on the other hand, in order to extract meta-knowledge during the meta-training process, graph meta-learning models will be trained with diverse node classification tasks from disjoint label spaces. It requires the model to quickly adapt to never-before-seen labels, which poses great challenges to existing denoising algorithms since they are optimized in a static learning environment (i.e., a single task). Hence, how to bridge the gap between the two learning paradigms and design a graph meta-learning model that can effectively denoise and efficiently adapt to new task (i.e., \textit{meta-test} task) with \textit{unseen classes} is vital to be explored.


To address the aforementioned challenges, we present a new robust graph meta-learning framework -- Graph Interpolation Network (Meta-GIN) in this paper. Meta-GIN is meat-learned with our carefully designed \textit{robustness-enhanced episodic training}, which can effectively solve weakly-supervised few-shot node classification on graphs. 
Instead of directly learning from a noisy meta-training task in each episode, Meta-GIN meta-learns the graph FSL model from the noise-reduced meta-training task interpolated across a set of meta-training tasks.
To obtain each noise-reduced node representation in an interpolated meta-training task, Meta-GIN randomly samples a set of meta-training tasks sharing the same label space and learn expressive node representations that captures both node attributes and topological structure information. Afterwards, by interpolating and comparing the information among the sampled nodes with the same label across a set of meta-training task, Meta-GIN is able to provide a better estimation of confidence score for each node and further summarize a noise-reduced representation of the corresponding target class. By learning across a pool of those interpolated tasks, Meta-GIN can be meta-learned not only to denoise from weakly-labeled data, but also to extrapolate the knowledge from seen to unseen node classes. Finally, the learned Meta-GIN can quickly adapt to new tasks using few fine-tuning steps.  In summary, the main contributions of our work are: 


\begin{itemize}[leftmargin=*,noitemsep,topsep=1.5pt]

\item We investigate a new problem -- weakly-supervised graph meta-learning, which can mitigate the limitation of existing graph meta-learning methods and push forward the frontiers of few-shot learning on graphs.
    
\item  We propose a principled framework Meta-GIN that is capable of extracting highly transferable meta-knowledge from weakly-labeled data, in order to solve unseen node classification tasks with few labeled nodes. 
    
\item  We perform extensive experiments on various real-world datasets to corroborate the effectiveness of our approach. The experimental results demonstrate the superior performance of Meta-GIN over existing efforts.

\end{itemize}

%% file: Related.tex
 Graph neural networks~\cite{chang2015heterogeneous,cao2016deep,kipf2017semi,velickovic2017graph, hamilton2017inductive} have recently achieved momentous success in transforming the information of a graph into low-dimensional latent representations. The effectiveness of prevailing graph ML methods such as GNNs largely relies on sufficient labeled instances. However, those methods fail to address graph few-shot learning (FSL) problems, where the unseen concepts during testing phrase only have few labeled instances~\cite{zhou2019meta}. 

Meta-learning, also known as learning-to-learn, enables the models to accumulate knowledge from previous experiences, and has led to significant progress in various domains for addressing FSL problems. In essence, a meta-learning model learns across diverse \textit{meta-training} tasks sampled from those \textit{seen classes} with a large quantity of labeled data, and can be naturally generalized to a new task (i.e., \textit{meta-test} task) with \textit{unseen classes} during training~\cite{snell2017prototypical}. Following this learning paradigm, researchers have proposed to use GNNs as the backbone to extrapolate meta-knowledge on graphs, which demonstrates promising results. Among recent advances of FSL on graphs, a major line of work aims to solve the task of node classification~\cite{zhou2019meta,ding2020graph,huang2020graph,liu2021relative,lan2020node,liu2020towards}. Among them, Meta-GNN, GPN and G-Meta are three representative ones. Specifically, Meta-GNN~\cite{zhou2019meta} uses gradient-based meta-learning to optimize a GNN model for few-shot node classification. GPN~\cite{ding2020graph} extends prototypical networks to graph-structured data by considering the importance of each node. G-Meta~\cite{huang2020graph} uses local subgraphs to transfer subgraph-specific information. Note that other methods such as GFL~\cite{yao2019graph} and MetaTNE~\cite{lan2020node} are different from our scenario since they are focusing on multiple networks and plain networks, respectively. In addition to few-shot node classification, other graph ML tasks including graph classification~\cite{chauhan2019few,ma2020adaptive} and link prediction~\cite{baek2020learning,zhang2020few} has also been studied under the FSL setting. Unlike previous works, we propose a weakly-supervised graph meta-learning framework, which eliminates the dependency of gold-labeled data during meta-training.



%% file: Methodology.tex

In this section, we introduce the details of the proposed Meta-GIN. To better explain how it works, we show its framework in Figure 1. Based on our robustness-enhanced episodic training, Meta-GIN facilitates graph meta-learning on weakly-labeled nodes by interpolated noise-reduced node representations. With the noise-reduced node representations, Meta-GIN further extracts highly transferable meta-knowledge and performs few-shot node classification on novel classes using optimization-based meta-learning. In
the following subsections, we elaborate three key parts: robustness-enhanced episodic training, graph interpolation networks, and meta-optimization, respectively.

\begin{problem}
\textbf{Weakly-supervised Few-shot Node Classification}:
Given an attributed graph $\mathbf{G}=(\mathbf{A}, \mathbf{X})$,
and the node label set $\mathcal{Y} = \{y_1, y_2, . . . , y_c\}$ that can be divided two subsets: the seen labels $\mathcal{Y}_{train}$, and the unseen labels $\mathcal{Y}_{test}$. Specifically, we have substantial weakly-labeled nodes with label noise for $\mathcal{Y}_{train}$, and few-shot clean-labeled nodes (i.e., support set $\mathcal{S}$) for each class in $\mathcal{Y}_{test}$. The problem aims to study how to predict labels for the unlabeled nodes (i.e., query set $\mathcal{Q}$) from those few-shot node classes $\mathcal{Y}_{test}$, by leveraging the knowledge of weakly-labeled nodes from $\mathcal{Y}_{train}$. 
\end{problem}


Note that if $\mathcal{Y}_{test}$ consists of $N$ classes and the support set $\mathcal{S}$ includes $K$ labeled nodes per class, this problem is named $N$-way $K$-shot node classification problem. In essence, the objective of this problem is to learn a meta-classifier that can be adapted to new classes with only a few labeled nodes. Therefore, how to extract highly transferable meta-knowledge from weakly-labeled data from  $\mathcal{Y}_{train}$ is the key for solving the studied problem.

\subsection{Robustness-enhanced Episodic Training}
\label{sec:episodic}



The effectiveness of few-shot learning algorithms largely benefits from the episodic training paradigm~\cite{vinyals2016matching}. Briefly, the key idea of episodic training is to mimic the real test environment by sampling data from $\mathcal{Y}_{train}$ and the model learns over such \textit{meta-training} tasks in a large number of episodes. Following this idea, graph FSL methods construct a pool of few-shot node classification tasks according
to the seen labels. For each meta-training task $\mathcal{T}_t = \{\mathcal{S}_t, \mathcal{Q}_t\}$, the model is trained to minimize the loss of its predictions for the query set $\mathcal{Q}_t$ , and goes episode by episode until convergence. In this way, the model gradually collects meta-knowledge across those meta-training tasks and then can be naturally generalized to the meta-testing task $\mathcal{T}_{test} = \{\mathcal{S}, \mathcal{Q}\}$ with unseen classes $\mathcal{Y}_{test}$.


 \begin{figure*}[t]
  \vspace{-0.15in}
    \graphicspath{{figures/}}
    \centering
    \includegraphics[width=0.95\textwidth]{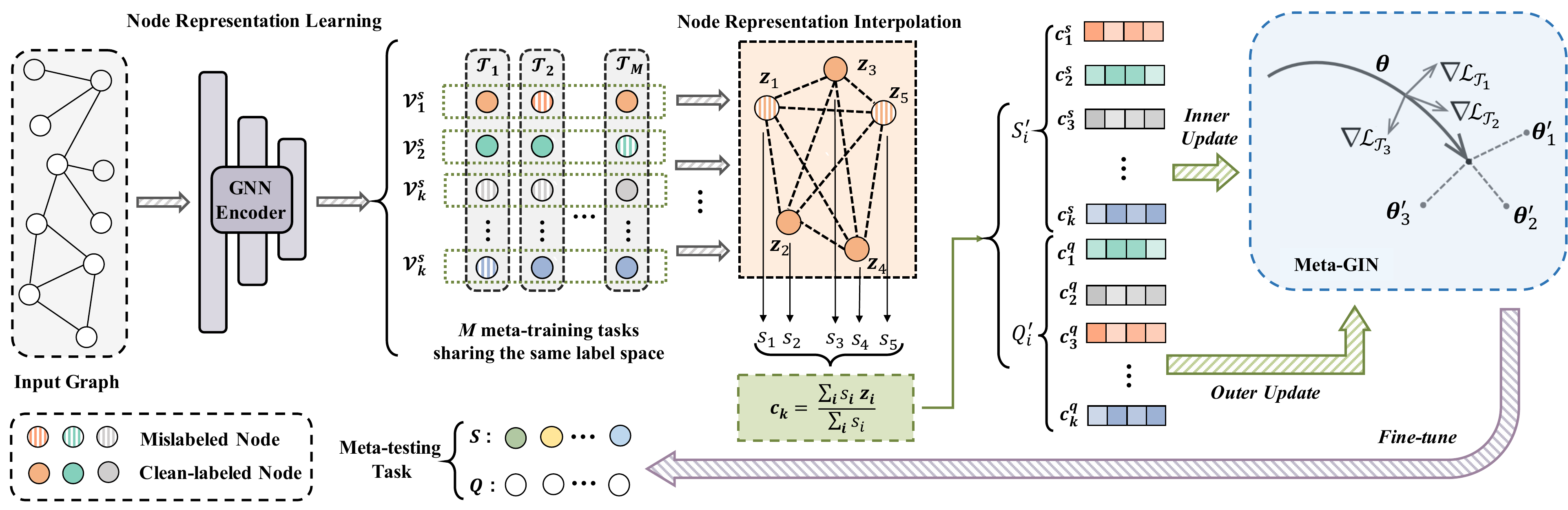}
    \caption{Overview of the proposed framework. In each training episode, Meta-GIN interpolates weakly-labeled nodes from multiple meta-training tasks to obtain the noise-reduced support and query set and further extracts highly transferable meta-knowledge. During meta-testing, Meta-GIN can quickly adapt to unseen tasks with few-shot labeled instances.}%
    \label{fig:framework}%
\end{figure*}

\smallskip
\noindent\textbf{Meta-training with Task Interpolation.} However, existing graph FSL methods commonly assume that the labels of nodes from $\mathcal{Y}_{train}$ are clean, which is invalid under the weakly-supervised setting we target. To suppress the label noise during the meta-training process, we propose a robustness-enhanced episodic training paradigm by using the idea of \textit{task interpolation}. Specifically, we sample $M$ meta-training tasks that share the same label space and then perform interpolation among the $M$ tasks to generate a \textit{noise-reduced meta-training task}. Correspondingly, each node in the final $N$-way $K$-shot task is interpolated by $M$ nodes from different tasks with the same class label. 

During each episode training, we firstly sample a set of $M$ meta-training tasks $\{\mathcal{T}_t\}_{t=1}^M$ sharing the same label space from $\mathcal{Y}_{train}$:
\begin{equation}
    \begin{aligned}
        \mathcal{S}_t &= \{(v_{t,1}^s, y_{t,1}^s),(v_{t,2}^s, y_{t,2}^s), . . . ,(v_{t, N \times K}^s, y_{t, N \times K}^s)\},\\
        \mathcal{Q}_t &= \{(v_{t,1}^q, y_{t,1}^q),(v_{t,2}^q, y_{t,2}^q), . . . ,(v_{t, N \times K'}^q, y_{t, N \times K'}^q)\},\\
        \{\mathcal{T}_t\}_{t=1}^M &= \{\{\mathcal{S}_1, \mathcal{Q}_1\}, \{\mathcal{S}_2, \mathcal{Q}_2\}, ..., \{\mathcal{S}_M, \mathcal{Q}_M\}\},
    \end{aligned}
\end{equation}
where all the task in $\{\mathcal{T}_t\}_{t=1}^M$ are sampled from the same $N$ classes. For each meta-training task $\mathcal{T}_t$, the support set $\mathcal{S}_t$ contains $N$ classes and $K$ weakly-labeled nodes per class, while $\mathcal{Q}_t$ containing $K'$ query nodes sampled from the remainder of each of the $N$ classes.

Furthermore, for each set of $M$ label-sharing meta-training tasks with weakly-labeled nodes, the proposed framework Meta-GIN will try to generate a noise-reduced meta-training task via task interpolation to improve the effectiveness of episodic training under the weakly-supervised setting. Thus we can get the noise-reduced support and query sets:
\begin{equation}
    \begin{aligned}
        \mathcal{S}' &= \{ (\mathbf{c}_1^s, y_1^s) , (\mathbf{c}_2^s, y_2^s), . . . ,(\mathbf{c}_{N \times K}^s, y_{N \times K}^s)\},\\
        \mathcal{Q}' &= \{(\mathbf{c}_1^q, y_1^q) , (\mathbf{c}_2^q, y_2^q), . . . ,(\mathbf{c}_{N \times K'}^q, y_{N \times K'}^q)\},
    \end{aligned}
\end{equation}
where $\mathbf{c}_k$ denotes the noise-reduced node representation generated from $\mathcal{V}_k = \{v_{1,k}, v_{2,k}, ..., v_{M,k}\}$ across the $M$ tasks and $y_k$ denotes its corresponding shared target class. With the noise-reduced meta-training task, our model could further extract highly transferable meta-knowledge and solve the weakly-labeled graph meta-learning problems.

\subsection{Graph Interpolation Networks}
Moreover, we propose a new family of graph neural networks, called Graph Interpolation Networks (GIN) to facilitate graph meta-learning on weakly-labeled nodes. In essence, GIN is composed of two key building blocks, including (1) a \textit{node representation learning} module that embeds each node; and (2) a \textit{node interpolation} module for deriving noise-reduced node representations for the final noise-reduced meta-training task. The details are as follows:

\smallskip
\noindent\textbf{Node Representation Learning.} The initial step of conducting graph meta-learning is to learn expressive node representations that capture both graph structure and node features. To achieve this, we first design a GNN-based encoding module in GIN. Specifically, it is built with multiple GNN layers that encode each node to a low-dimensional latent representation. The core operation in GNNs is the message passing scheme, in which information is propagated from each node to its neighborhoods with specific deterministic propagation rules. 
It is worth noting that the encoding module is compatible with arbitrary GNN-based architecture~\cite{kipf2017semi,hamilton2017inductive,velickovic2017graph}. To improve the model efficiency on large graphs, we employ Simple Graph Convolution (SGC)~\cite{wu2019simplifying} in this work. Specifically, SGC utilizes a simplified graph convolution pre-processing followed by standard multi-class logistic regression. 

\smallskip
\noindent\textbf{Node Representation Interpolation.} 
However, the previously learned node representations from seen classes $\mathcal{Y}_{train}$ are unable to represent their corresponding classes since they are likely mislabeled. Hence, our \textit{node representation interpolation} module aims at interpolating nodes in node set $\mathcal{V}_k$ to generate noise-reduced node representations and leverage those interpolated node representations to learn the concept of each class.

To generate a noise-reduced representation with the $M$ labeled nodes from a set $\mathcal{V}_k$, one straightforward solution is taking the average of all the embedded nodes belonging to that set with $\mathbf{p}_k = \frac{1}{|\mathcal{V}_k|} \sum_{i \in \mathcal{V}_k} \mathbf{z}_i$,
where $\mathbf{z}_i$ is the learned node representations from the \textit{graph representation learning} module of node $v_i$. However, directly taking the mean vectors of the embedded nodes as noise-reduced node representation could be ineffective due to the existence of mislabeled nodes. Specifically, our \textit{node representation interpolation} module is designed to estimate a confidence score $\alpha_i$ for each node and further perform fine-grained interpolation among the nodes from $M$ tasks.





To identify the confidence score of each labeled node, GIN will compute the confidence score of each sample with message passing. As shown in Figure~\ref{fig:framework}, we first build a full-connected interpolation graph using $M$ sampled nodes from each set $\mathcal{V}_k$, then we develop a node re-weighting layer using graph attention mechanism to aggregate and compare the information among weakly-labeled nodes. The node re-weighting layer can be defined as follows:
\begin{equation}
    s_i =  \sigma\bigg(\sum_{v_j \in \mathcal{V}_{k}} \alpha_{ij}   \mathbf{w}^{\mathrm{T}} [\mathbf{z}_j || \bm{\Delta}_{j}] \bigg),
    \label{eq:weight}
\end{equation}
where $\mathbf{w} \in \mathbb{R}^{2 d'}$ is the learnable parameter vector, $\sigma$ is a nonlinear activation function, i.e., sigmoid function. $s_i$ denote the confidence score of node $v_i$ and $\Delta_{j} = \mathbf{z}_j - \mathbf{p}_k$ captures the difference between the embedding of node $v_j$ and the prototype of $\mathcal{V}_k$. By incorporating the distance between each node and the prototype, GIN can better perceive the concept of the corresponding class and compute the final confidence score. Specifically, $\alpha_{ij}$ is the attention weight between nodes $v_i$ and $v_j$, we compute it via attention:
\begin{equation}
    \alpha_{ij} = \frac{\text{exp}\big(\text{LeakyReLU}\big(\mathbf{a}^{\mathrm{T}}[\mathbf{w}^{\mathrm{T}} \tilde{\mathbf{z}}_i || \mathbf{w}^{\mathrm{T}}  \tilde{\mathbf{z}}_j]\big)\big)}{\sum_{m \in \mathcal{V}_k} \text{exp}\big(\text{LeakyReLU}\big(\mathbf{a}^{\mathrm{T}}[\mathbf{w}^{\mathrm{T}}  \tilde{\mathbf{z}}_i || \mathbf{w}^{\mathrm{T}}  \tilde{\mathbf{z}}_m]\big)\big)},
\end{equation}
where $\tilde{\mathbf{z}}_i = [\mathbf{z}_i || \bm{\Delta}_i]$ and the attention vector $\mathbf{a}$ is a trainable weight vector that assigns importance to different node during aggregation.

After obtaining the confidence scores of the weakly-labeled nodes in each set $\mathcal{V}_k$, GIN is able to generate a noise reduced node representation by interpolating these noisy labeled nodes with their weights. Based on the computed attentional weights from Eq. (\ref{eq:weight}), we can obtain the interpolated representation with $\mathbf{c}_k = \frac{1}{\sum_i  s_i}{\sum_i s_i \mathbf{z_i}}$.




\smallskip
\noindent\textbf{Node Classification.} With the noise-reduced support set $\mathcal{S}'$ that contains $K$ interpolated node representations for each of the $N$ classes, GIN will try to classify each instance to its corresponding class label. This can be done with a feed-forward layer $\mathbf{y}_k = \text{softmax}(\mathbf{W}^{\mathrm{T}}_c \mathbf{c}_k + \mathbf{b}_c)$,
where $\mathbf{W}_c \in \mathbb{R}^{d' \times N}$ and $\mathbf{b}_c \in \mathbb{R}^{N}$ are learnable weight matrix and bias, respectively. Under the episodic training framework, the objective of each meta-training task is to minimize the cross-entropy loss function for performing node classification. Specifically, the training loss for each interpolated instance $\mathbf{c}_k$ is computed by:
\begin{equation}
    \mathcal{L} = -  \log p(y_k^* | \mathbf{c}_k),
\label{eq:loss}
\end{equation}
where $y_k^*$ is the shared label of set $\mathcal{V}_k$. As the training instances are computed by the \textit{node interpolation} module, GIN is able to reduce the negative impacts of mislabeled nodes during the meta-learning process. By minimizing the above loss function, GIN is able to learn a generic classifier for a specific $N$-way $K$-set meta-training task and further extract highly transferrable meta-knowledge from weakly-labeled data.


\subsection{Meta-optimization}
\label{sec:meta-optimization}

Having the proposed Graph Interpolation Networks, we are able to obtain noise-reduced support set $\mathcal{S}'$ and query set $\mathcal{Q}'$ via interpolating multiple meta-training tasks. Upon that, we train the model via meta-learning, such that the meta-learned GIN model (Meta-GIN) is capable of effectively adapting to new tasks with few labeled instances. Specifically, we follow model-agnostic meta-learning~\cite{finn2017model} to learn Meta-GIN in an optimization-based fashion, in order to better exploit the clean-labeled support nodes during meta-testing and make fast and effective adaptation to a new task through a small number of gradient steps.


\smallskip
\noindent\textbf{Meta-training.} In the meta-training stage, we expect to obtain a good initialization of GIN, which is inherently generalizable to unseen tasks, and explicitly encourage the initialization parameters to perform well after a small number of gradient descent updates on a new learning task. When learning a specific interpolated task $\mathcal{T}_i'$, we begin with feeding the nodes from the noise-reduced support set $\mathcal{S}_i'$ to GIN, and calculate the cross-entropy loss $\mathcal{L}_{\mathcal{T}_i'}$ as formulated in Eq. (\ref{eq:loss}). We consider a GIN model represented by a parameterized function $f_{\bm\theta}$ with parameters $\bm\theta$, the optimization algorithm first adapts the initial model parameters $\bm\theta$ to $\bm\theta_i'$ for each interpolated learning task $\mathcal{T}_i'$ independently. Specifically, the updated parameter $\bm\theta_i'$ is computed using $\mathcal{L}_{\mathcal{T}_i'}$ on the interpolated node representation and the corresponding class label. Formally, the parameter update with one gradient step can be expressed as:
\begin{equation}
    \bm\theta_i' = \bm\theta - \alpha\nabla_{\bm\theta}\mathcal{L}_{\mathcal{T}_i'}(f_{\bm\theta}),
    \label{eqn:update_theta}
\end{equation}
where $\alpha$ controls the learning rate for each task. Note that Eq.~\eqref{eqn:update_theta} only includes one-step gradient update, while it is straightforward to extend to multiple gradient updates~\cite{finn2017model}.

\begin{algorithm}[t]
\caption{The learning algorithm of Meta-GIN.}
\label{alg:GIN}
\LinesNumbered
\small
\KwIn{Task distribution $p(\mathcal{T})$ over the input graph $\mathbf{G}$}
\KwOut{The well-trained model Meta-GIN}
Randomly initialize the parameters $\bm\theta$ of GIN

\While{not converge}{
    Randomly sample a batch of task sets with all the tasks in a set $\{\mathcal{T}_t\}_{t=1}^M$ sharing the same label space. 

    \For{each task set $\{\mathcal{T}_t\}_{t=1}^M$}{
    
    // \texttt{Node Interpolation}
    
    \For{each $\mathcal{V}_k \in \{\{\mathcal{S}_t\}_{t=1}^M, \{\mathcal{Q}_t\}_{t=1}^M\}$ }{
    Compute the representations for nodes in $\mathcal{V}_k$
    
    Interpolate and obtain the noise-reduced node representation $\mathbf{c}_k$ 
    }
    
    Obtain the noise-reduced $\mathcal{S}_i'$ and $\mathcal{Q}_i'$
    
    Evaluate $\nabla_{\bm\theta}\mathcal{L}_{\mathcal{T}_i'}(f_{\bm\theta})$ using $\mathcal{S}_i'$ and $\mathcal{L}_{\mathcal{T}_i'}$ in Eq. (\ref{eq:loss})

    Compute adapted parameters $\bm\theta'$ by Eq. (\ref{eqn:update_theta})

    }
    Update $\bm\theta$ with the interpolated query set by Eq. (\ref{eqn:meta_loss})
     
    }
    
    \Return Meta-learned graph meta-learning model Meta-GIN

\end{algorithm}

Our model Meta-GIN is trained by optimizing for the best performance of $f_{\bm\theta}$ with respect to $\bm\theta$ across all interpolated meta-training tasks. More concretely, the meta-objective function is defined as follows:
\begin{equation}
\min_{\bm\theta} \sum_{\mathcal{T}_i' \sim p(\mathcal{T})} \mathcal{L}_{\mathcal{T}_i'}(f_{\bm\theta_i'}) = \min_{\bm\theta}\sum_{\mathcal{T}_i' \sim p(\mathcal{T})}\mathcal{L}_{\mathcal{T}_i'}(f_{\bm\theta - \alpha\nabla_{\bm\theta}\mathcal{L}_{\mathcal{T}_i'}(f_{\bm\theta})}),
    \label{eqn:meta_loss}
\end{equation}
where $p(\mathcal{T})$ is the distribution of interpolated meta-training tasks. Since the meta-optimization is performed over parameters $\bm\theta$ with the objective computed using the updated parameters (i.e., $\bm\theta_i'$) for all tasks, correspondingly, the model parameters are optimized such that one or a small number of gradient steps on the target task will produce maximal effectiveness.


Formally, we leverage stochastic gradient descent (SGD) to update the model parameters $\bm\theta$ with the instances from the interpolated query set, such that the model parameters $\bm\theta$ are updated as follows:
\begin{equation}
    \bm\theta \leftarrow \bm\theta - \beta\nabla_{\bm\theta}\sum_{\mathcal{T}_i' \sim p(\mathcal{T})} \mathcal{L}_{\mathcal{T}_i'}(f_{\bm\theta_i'}),
\end{equation}
where $\beta$ is the meta step size. The detailed learning process of Meta-GIN is presented in Algorithm \ref{alg:GIN}.

\smallskip
\noindent\textbf{Meta-testing.}
After training on a considerable number of meta-training tasks, we expect that the Meta-GIN model has been gradually meta-learned well for handling unseen few-shot node classification tasks. Its generalization performance will be measured on the test episodes, which contain clean-labeled nodes sampled from $\mathcal{Y}_{test}$ instead of $\mathcal{Y}_{train}$. For each meta-testing episode, we we will remove the Node Interpolation module, and fine-tune the meta-learned classifier Meta-GIN with the provided clean-labeled support set $\mathcal{S}$ and classify each query node in $\mathcal{Q}$ into the most likely class. 



%% file: Experiments.tex
In this section, we will start with the experimental setup and then present our experiment results to answer three research questions: (i) is Meta-GIN effective in solving the weakly-supervised few-shot node classification? (ii) Can Meta-GIN achieve satisfying performance on a variety of noise levels? and (iii) How does each module in Meta-GIN contribute to the final performance?

%

\begin{table*}[t!]

\centering

\caption{Test accuracy for weakly-supervised few-shot node classification (30\% label noise) on different datasets.}
\vspace{-0.05in}

\setlength{\tabcolsep}{2.5pt}
\scalebox{0.82}{
\begin{tabular}{@{}l cc c cc c cc c cc @{}}
\toprule

\multirow{2}{*}{\textbf{Amazon}} & \multicolumn{2}{c}{5-way 1-shot}  & &  \multicolumn{2}{c}{5-way 3-shot}  &  & \multicolumn{2}{c}{10-way 1-shot}  &  & \multicolumn{2}{c}{10-way 3-shot} 

\\ \cline{2-3} \cline{5-6} \cline{8-9} \cline{11-12}


\rule{0pt}{10pt} & \multicolumn{1}{c}{Symmetric}  & \multicolumn{1}{c}{Asymmetric } &&   \multicolumn{1}{c}{Symmetric}  & \multicolumn{1}{c}{Asymmetric} & &
\multicolumn{1}{c}{Symmetric}  & \multicolumn{1}{c}{Asymmetric} & &
\multicolumn{1}{c}{Symmetric}  & \multicolumn{1}{c}{Asymmetric}

\\ \midrule

GCN    & $0.386\pm0.025$  & $0.381\pm0.023$  & & $0.534\pm0.027$ & $0.529\pm0.023$  &&  $0.211\pm0.019$ & $0.197\pm0.021$ && $0.376\pm0.027$ & $0.372\pm0.022$   \\

SGC &  $0.390\pm0.019$  & $0.385\pm0.021$  & & $0.535\pm0.018$ & $0.527\pm0.017$  &&  $0.248\pm0.016$ & $0.232\pm0.012$ && $0.383\pm0.013$ & $0.380\pm0.015$   \\

GraphSAGE &   $0.338\pm0.018$  & $0.406\pm0.023$  & & $0.529\pm0.021$ & $0.568\pm0.016$  && $0.211\pm0.026$  & $0.224\pm0.018$ && $0.362\pm0.026$ & $0.373\pm0.014$ \\
PTA &  $0.405\pm0.023$  & $0.437\pm0.018$  & & $0.553\pm0.012$ & $0.581\pm0.022$  && $0.252\pm0.016$  & $0.255\pm0.015$ && $0.394\pm0.023$ & $0.409\pm0.018$ \\
\midrule

Meta-GNN   &  $0.403\pm0.021$  & $0.480\pm0.018$  & & $0.601\pm0.028$ & $0.650\pm0.024$  && $0.279\pm0.034$  & $0.325\pm0.031$ && $0.555\pm0.023$ & $0.563\pm0.029$ \\

GPN   &  $0.408\pm0.015$  & $0.463\pm0.023$  & & $0.629\pm0.024$ & $0.651\pm0.027$  && $0.263\pm0.011$  & $0.314\pm0.026$ && $0.535\pm0.011$ & $0.579\pm0.016$ \\

G-Meta &  $0.488\pm0.025$  & $0.496\pm0.024$  & & $0.614\pm0.018$ & $0.658\pm0.022$  && $0.336\pm0.025$  & $0.376\pm0.021$ && $0.463\pm0.021$ & $0.504\pm0.019$  \\

Mate-GPS &  $0.495\pm0.021$  & $0.512\pm0.019$  & & $0.631\pm0.019$ & $0.662\pm0.025$  && $0.359\pm0.018$  & $0.397\pm0.019$ && $0.541\pm0.024$ & $0.561\pm0.021$  \\



\textbf{Meta-GIN}   &  $\mathbf{0.694\pm0.033}$  & $\mathbf{0.652\pm0.010}$  & & $\mathbf{0.750\pm0.021}$ & $\mathbf{0.725\pm0.021}$  && $\mathbf{0.567\pm0.027}$  & $\mathbf{0.559\pm0.025}$ && $\mathbf{0.627\pm0.028}$ & $\mathbf{0.615\pm0.022}$ \\

\bottomrule
\end{tabular}}

\medskip
\setlength{\tabcolsep}{2.5pt}
\scalebox{0.82}{
\begin{tabular}{@{}l cc c cc c cc c cc @{}}
\toprule

\multirow{2}{*}{\textbf{DBLP}} & \multicolumn{2}{c}{5-way 1-shot}  & &  \multicolumn{2}{c}{5-way 3-shot}  &  & \multicolumn{2}{c}{10-way 1-shot}  &  & \multicolumn{2}{c}{10-way 3-shot} 

\\ \cline{2-3} \cline{5-6} \cline{8-9} \cline{11-12}


\rule{0pt}{10pt} & \multicolumn{1}{c}{Symmetric}  & \multicolumn{1}{c}{Asymmetric} &&   \multicolumn{1}{c}{Symmetric}  & \multicolumn{1}{c}{Asymmetric} & &
\multicolumn{1}{c}{Symmetric}  & \multicolumn{1}{c}{Asymmetric} & &
\multicolumn{1}{c}{Symmetric}  & \multicolumn{1}{c}{Asymmetric}

\\ \midrule

GCN    & $0.369\pm0.022$  & $0.345\pm0.031$  & & $0.470\pm0.041$ & $0.444\pm0.031$  &&  $0.212\pm0.030$ & $0.201\pm0.014$ && $0.345\pm0.033$ & $0.335\pm0.042$   \\

SGC &  $0.376\pm0.023$  & $0.374\pm0.016$  & & $0.487\pm0.024$ & $0.479\pm0.023$  &&  $0.224\pm0.013$ & $0.223\pm0.014$ && $0.359\pm0.019$ & $0.348\pm0.025$   \\

GraphSAGE &  $0.345\pm0.021$  & $0.354\pm0.028$  & & $0.536\pm0.021$ & $0.540\pm0.024$  && $0.262\pm0.014$  & $0.279\pm0.015$ && $0.350\pm0.018$ & $0.395\pm0.023$ \\

PTA &  $0.388\pm0.020$  & $0.392\pm0.013$  & & $0.550\pm0.024$ & $0.561\pm0.025$  && $0.280\pm0.016$  & $0.303\pm0.019$ && $0.366\pm0.022$ & $0.427\pm0.025$ \\

\midrule

Meta-GNN & $0.581\pm0.010$  & $0.611\pm0.009$  & & $0.684\pm0.021$ & $0.702\pm0.025$  && $0.498\pm0.028$  & $0.514\pm0.025$ && $0.573\pm0.021$ & $0.578\pm0.024$   \\

GPN   & $0.566\pm0.020$  & $0.621\pm0.014$  & & $0.758\pm0.009$ & $0.766\pm0.014$  && $0.464\pm0.011$  & $0.501\pm0.018$ && $0.649\pm0.017$ & $0.650\pm0.015$  \\

G-Meta & $0.618\pm0.027$  & $0.627\pm0.022$  & & $0.697\pm0.025$ & $0.761\pm0.010$  && $0.497\pm0.025$  & $0.502\pm0.014$ && $0.536\pm0.027$ & $0.605\pm0.012$      \\

Mate-GPS &  $0.631\pm0.018$  & $0.639\pm0.020$  & & $0.749\pm0.016$ & $0.759\pm0.013$  && $0.504\pm0.021$  & $0.511\pm0.017$ && $0.613\pm0.018$ & $0.629\pm0.020$  \\

\textbf{Meta-GIN}  & $\mathbf{0.734\pm0.012}$  & $\mathbf{0.739\pm0.019}$  & & $\mathbf{0.794\pm0.010}$ & $\mathbf{0.772\pm0.016}$  && $\mathbf{0.591\pm0.023}$  & $\mathbf{0.593\pm0.020}$ && $\mathbf{0.672\pm0.020}$ & $\mathbf{0.695\pm0.011}$ \\

\bottomrule
\end{tabular}}

\medskip
\setlength{\tabcolsep}{2.5pt}
\scalebox{0.82}{
\begin{tabular}{@{}l cc c cc c cc c cc @{}}
\toprule

\multirow{2}{*}{\textbf{ogbn-}} & \multicolumn{2}{c}{5-way 1-shot}  & &  \multicolumn{2}{c}{5-way 3-shot}  &  & \multicolumn{2}{c}{10-way 1-shot}  &  & \multicolumn{2}{c}{10-way 3-shot} 

\\ \cline{2-3} \cline{5-6} \cline{8-9} \cline{11-12}


\textbf{arxiv}\rule{0pt}{10pt} & \multicolumn{1}{c}{Symmetric}  & \multicolumn{1}{c}{Asymmetric} &&   \multicolumn{1}{c}{Symmetric}  & \multicolumn{1}{c}{Asymmetric} & &
\multicolumn{1}{c}{Symmetric}  & \multicolumn{1}{c}{Asymmetric} & &
\multicolumn{1}{c}{Symmetric}  & \multicolumn{1}{c}{Asymmetric}

\\ \midrule

GCN    &   $0.259\pm0.033$  & $0.244\pm0.018$  & & $0.296\pm0.029$ & $0.288\pm0.021$  &&  $0.142\pm0.017$ & $0.122\pm0.015$ && $0.171\pm0.010$ & $0.157\pm0.011$   \\

SGC &   $0.277\pm0.020$  & $0.274\pm0.011$  & & $0.334\pm0.018$ & $0.321\pm0.009$  &&  $0.157\pm0.007$ & $0.155\pm0.011$ && $0.217\pm0.009$ & $0.196\pm0.012$   \\

GraphSAGE & $0.280\pm0.021$  & $0.260\pm0.019$  & & $0.317\pm0.022$ & $0.297\pm0.025$  &&  $0.143\pm0.016$ & $0.121\pm0.012$ && $0.170\pm0.015$ & $0.144\pm0.019$ \\
PTA &  $0.311\pm0.018$  & $0.303\pm0.020$  & & $0.352\pm0.016$ & $0.344\pm0.022$  && $0.187\pm0.017$  & $0.180\pm0.017$ && $0.235\pm0.014$ & $0.227\pm0.016$ \\

\midrule

Meta-GNN &  $0.451\pm0.017$  & $0.443\pm0.009$  & & $0.481\pm0.028$ & $0.478\pm0.026$  &&  $0.230\pm0.030$ & $0.222\pm0.018$ && $0.327\pm0.021$ & $0.302\pm0.011$  \\

GPN   &  $0.376\pm0.015$  & $0.420\pm0.019$  & & $0.492\pm0.018$ & $0.514\pm0.019$  &&  $0.255\pm0.019$ & $0.266\pm0.017$ && $0.266\pm0.015$ & $0.338\pm0.009$ \\

G-Meta &   $0.418\pm0.012$  & $0.422\pm0.014$  & & $0.453\pm0.013$ & $0.500\pm0.015$  &&  $0.272\pm0.012$ & $0.282\pm0.018$ && $0.355\pm0.017$ & $0.377\pm0.011$     \\

Mate-GPS &  $0.432\pm0.016$  & $0.438\pm0.015$  & & $0.469\pm0.013$ & $0.487\pm0.011$  && $0.263\pm0.011$  & $0.271\pm0.014$ && $0.335\pm0.015$ & $0.357\pm0.014$  \\

\textbf{Meta-GIN} &  $\mathbf{0.494\pm0.021}$  & $\mathbf{0.475\pm0.018}$  & & $\mathbf{0.572\pm0.013}$ & $\mathbf{0.545\pm0.016}$  &&  $\mathbf{0.336\pm0.021}$ & $\mathbf{0.325\pm0.023}$ && $\mathbf{0.447\pm0.013}$ & $\mathbf{0.390\pm0.013}$ \\

\bottomrule
\end{tabular}}
\label{table:semi}
\end{table*}

\subsection{Experiment Settings}
\label{sec:setting}
\smallskip
 \noindent{\textbf{Evaluation Datasets.}} 
 We adopt three datasets used in previous research~\cite{ding2019deep,huang2020graph} for few-shot node classification. 
  \textbf{Amazon}~\cite{mcauley2015inferring} is built with the products in ``Electronics'' and their complementary relationship (``bought together'') on Amazon. \textbf{DBLP}~\cite{tang2008arnetminer} is a DBLP citation network in which nodes denoting papers and the citation relations among papers are used to create links. To further compare the performance of different methods on large-scale graphs, we also include another citation network \textbf{ogbn-arxiv}, which is a benchmark dataset from Open Graph Benchmark (OGB). We follow the same train/validation/test splits and data preprocess procedure as in \cite{ding2020graph} for Amazon and DBLP datasets. For ogbn-arxiv dataset, we retrieve the network with the public OGB package and split it for few-shot learning node classification scenario.
  More details including the data sources, preprocessing procedures and the summary of statistics can be found in Appendix.

\smallskip
\noindent{\textbf{Label Corruption.}} To explore the performance of different methods for graph meta-learning on weakly-labeled data, we follow previous work~\cite{ren2018learning, chen2019understanding, jiang2018mentornet, hendrycks2018using} and inject two representative types of label noise to the datasets. Specifically, for a dataset with $P$ classes, \textit{\textbf{Symmetric Noise}} (Sym) corrupts each label class uniformly to all the other classes with probability $\epsilon/(P-1)$; and \textit{ \textbf{Asymmetric Noise}} (Asym) flips a label to a different class with probability $\epsilon$. To make the evaluation more realistic, \textit{both training and validation data will be perturbed}. More details can be found in Appendix.

\smallskip
\noindent{\textbf{Compared Methods.}}
In the experiments, we compare the proposed model GPN with two different categories of methods: (1) \textit{GNN-based methods}. \textbf{GCN}~\cite{kipf2017semi}, \textbf{SGC}~\cite{wu2019simplifying} and \textbf{GraphSAGE}~\cite{hamilton2017inductive} are three representative semi-supervised node classification methods. They are adapted for few-shot learning scenarios as in~\cite{zhou2019meta, ding2020graph}. \textbf{PTA}~\cite{dong2021equivalence} is a decoupled GNN which is robust to label noise using adaptive weighting strategy. (2) \textit{Graph meta-learning methods}. We include three state-of-the-arts for graph few-shot node classification: \textbf{Meta-GNN}~\cite{zhou2019meta} applies MAML~\cite{finn2017model} to SGC for few-shot node classification in graphs, while \textbf{GPN}~\cite{ding2020graph} is empowered by the Graph Prototypical Network to learn highly representative class prototypes and \textbf{Meta-GPS}~\cite{liu2022few} extends GPN with Prototype and Scaling \& shifting transformation. \textbf{G-Meta}~\cite{huang2020graph} regards the centroid embedding of local subgraphs as the prototypes and is optimized with both the prototypical loss and MAML.  

\begin{figure*}[!t]
\vspace{-0.05in}
    \graphicspath{{figures/}}
    \centering
    \scalebox{0.975}{
    \subfigure[5-w 1-s (Sym)]
    {
    \hspace{-0.2cm}
    \includegraphics[width=0.23\textwidth]{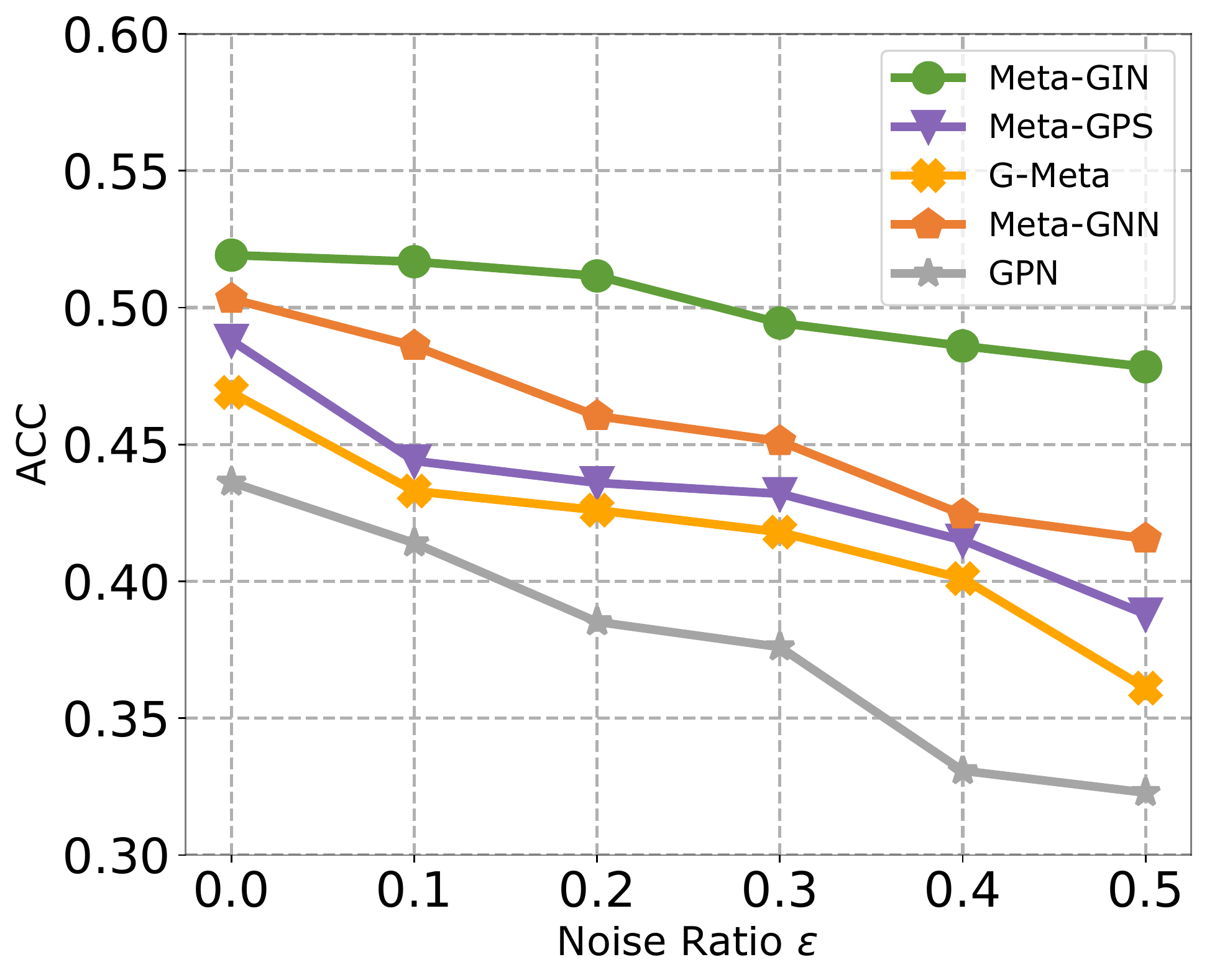}
    }
    \hspace{-0.2cm}
    \subfigure[5-w 1-s (Asym)]
    {
    \includegraphics[width=0.23\textwidth]{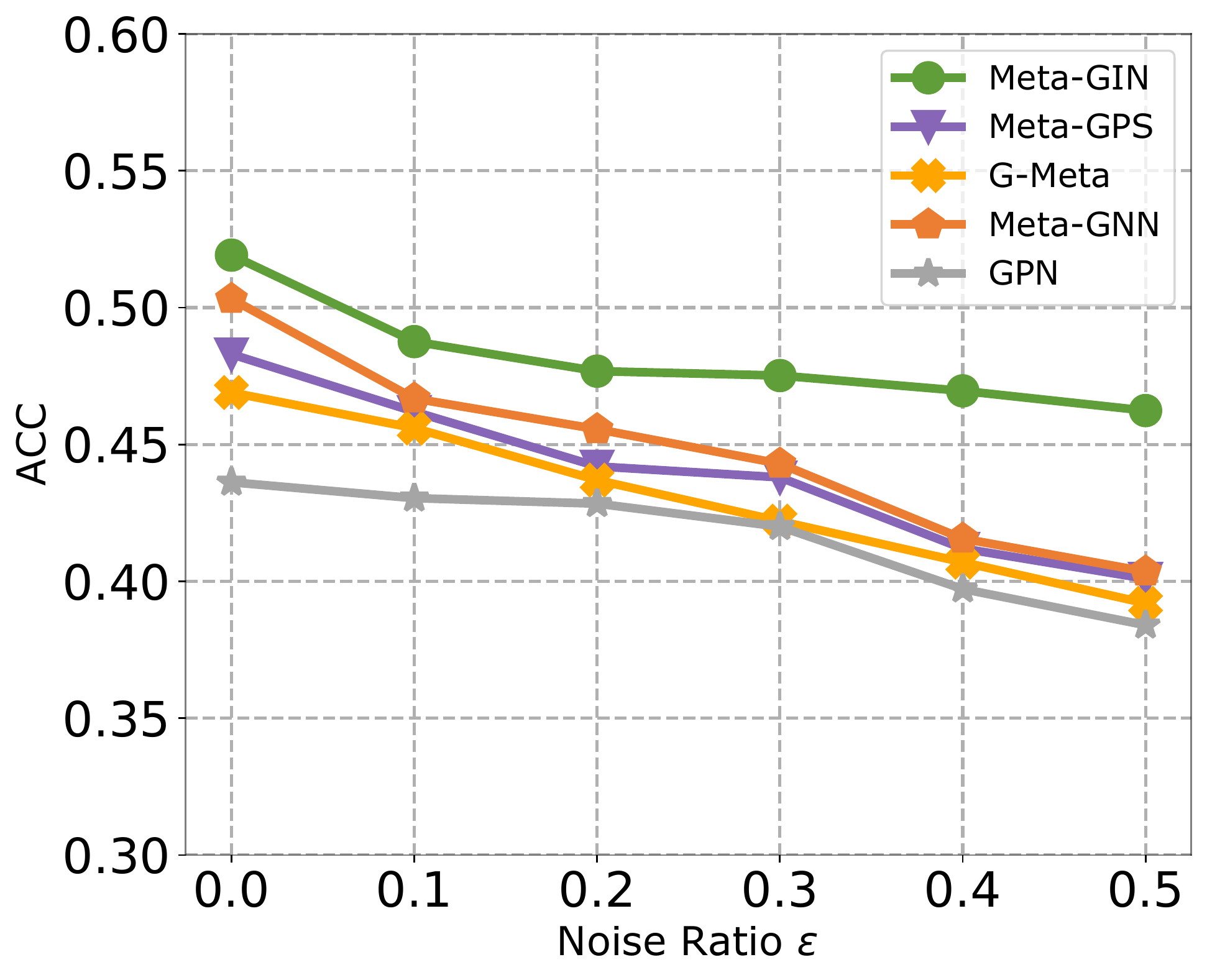}
    }
    \hspace{-0.2cm}
    \subfigure[10-w 1-s (Sym)] 
    {
    \includegraphics[width=0.23\textwidth]{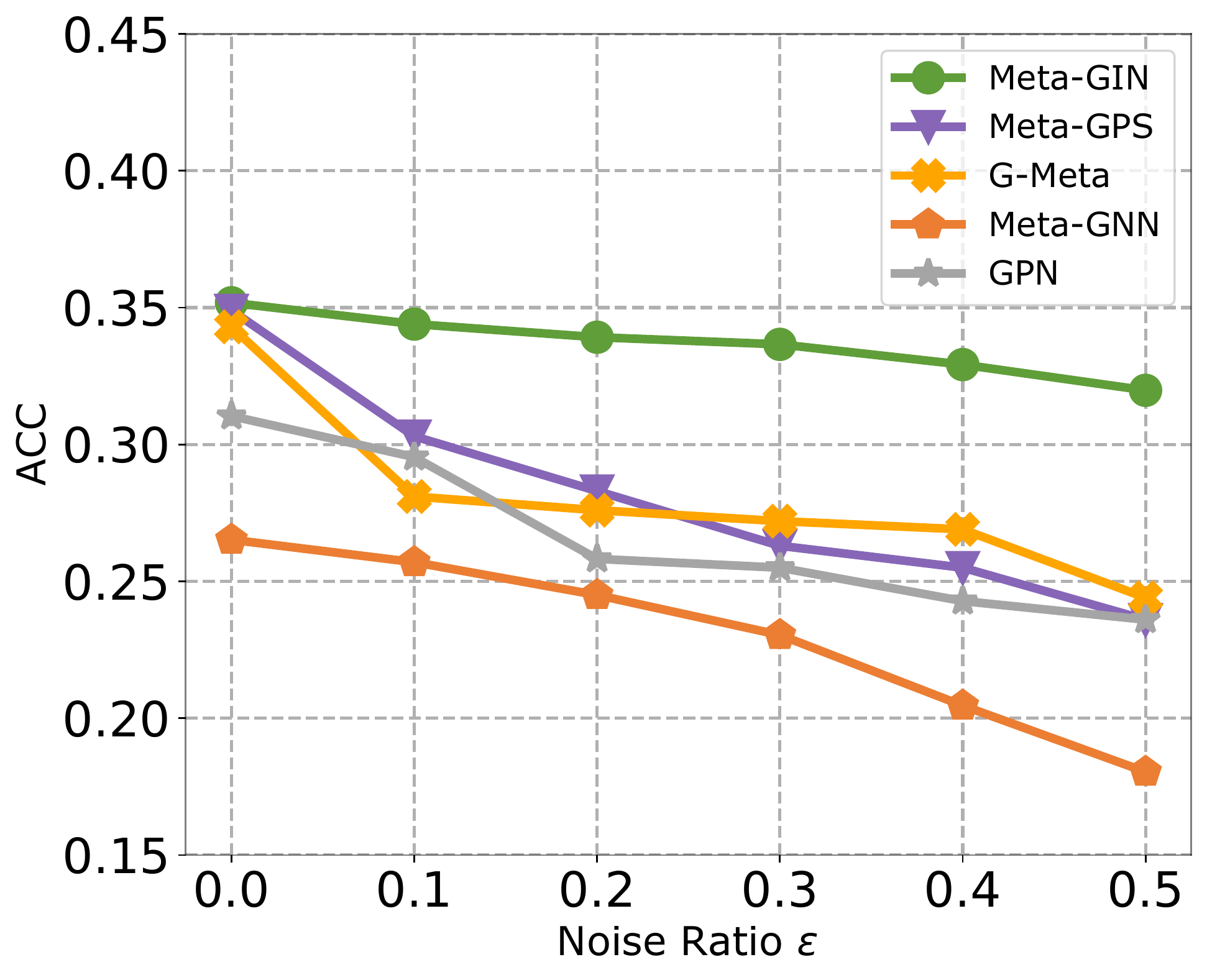}
    }
    \hspace{-0.2cm}
    \subfigure[10-w 1-s (Asym)]
    {
    \includegraphics[width=0.23\textwidth]{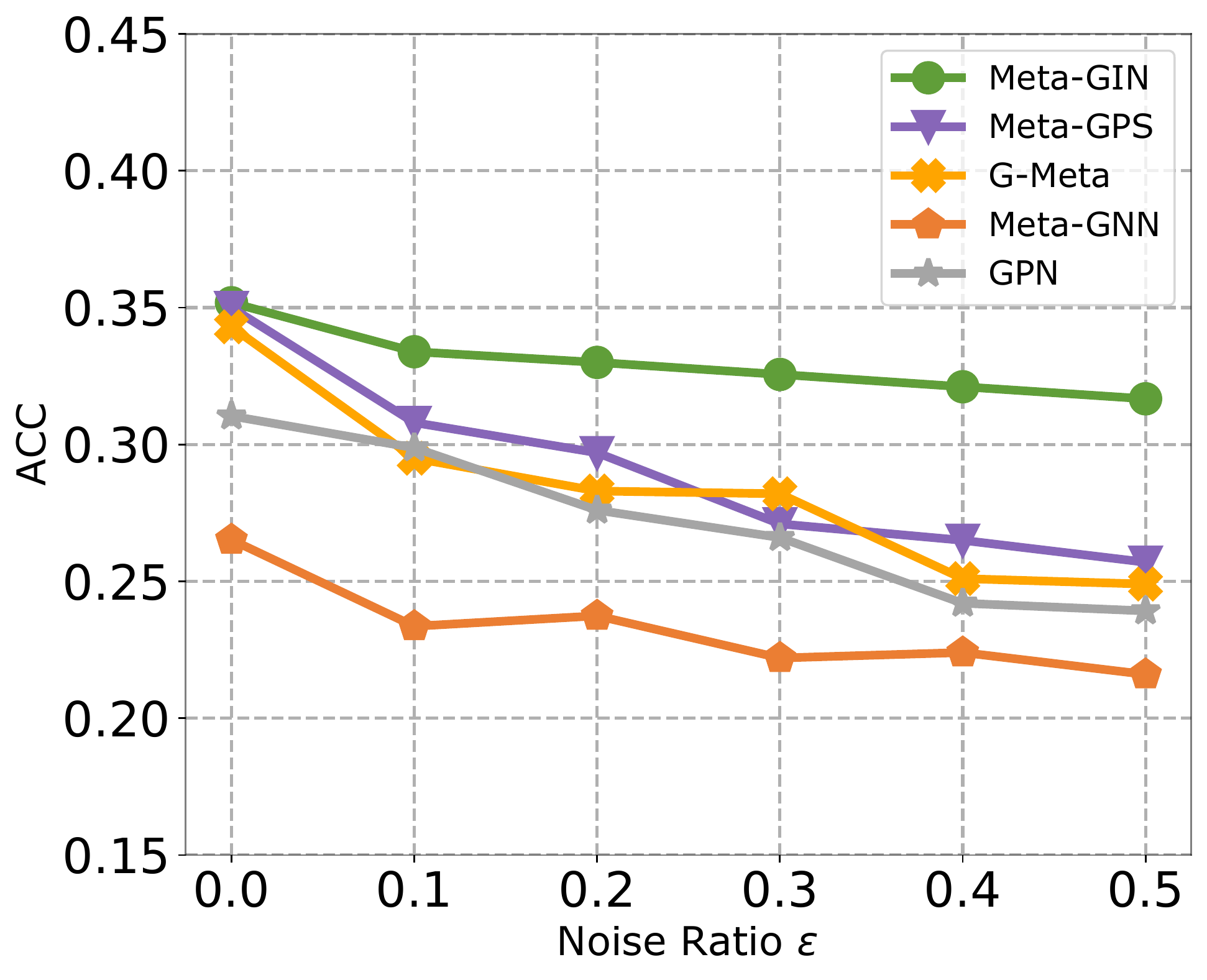}
    }}
     \vspace{-0.1in}
    \caption{Performance comparisons \textit{w.r.t.} different noise ratios on ogbn-arxiv dataset.}%
    \label{fig:N}
\end{figure*}


\subsection{Evaluation Results}

\smallskip
\noindent{\textbf{General Comparisons.}}
We evaluate the proposed Meta-GIN and all the baseline models on different weakly-supervised node classification tasks.
For each of the datasets, we inject either the symmetric noise or the asymmetric noise with noise ratio $\epsilon = 30\%$. Following previous work of few-shot node classification, we adopt \textit{Accuracy} (ACC) as the evaluation metric. We test each model 10 times and report the mean $\pm$ standard deviation in Table \ref{table:semi}. We can observe that the proposed Meta-GIN significantly outperforms all the baseline methods on weakly-supervised node classification tasks for different datasets corrupted by either symmetric or asymmetric label noise.

GNN-based methods such as GCN, SGC and GraphSAGE obtain poor performance while adapted for weekly-supervised few-shot learning scenarios since they require abundant clean labeled data to achieve satisfying classification accuracy. Though PTA can mitigate label noise to some extent, it is unable to transfer the knowledge from seen classes to unseen classes. For the graph FSL methods, they still fall behind the proposed Meta-GIN on the weakly-labeled data since they are vulnerable to label noise. Powered by the well-designed Graph Interpolation Networks under the robustness-enhanced episodic training framework, Meta-GIN is able to generate the noise-reduced node representation and achieve the best few-shot node classification performance on the noisy label data. It is also worth noting that noise usually have larger impact on meta-learning relying on fewer shot. However, compared with the graph meta-learning methods, the improvement of Meta-GIN on $N$-way-$1$-shot tasks is larger than that on $N$-way-$3$-shot tasks. This illustrates Meta-GIN's power on denoising for the practical few-shot learning scenarios.

\smallskip
\noindent{\textbf{Robustness Analysis.}}
To examine the robustness of Meta-GIN on data with different noise levels, we show its performance in Figure \ref{fig:N} by varying the noise ratio. 
Firstly, on the data with no injected noise (i.e., $\epsilon = 0$), Meta-GIN can still outperform the state-of-the-art for graph few-shot learning, which shows it is powerful in extrapolating the knowledge from seen to unseen node classes. Then if we inject the noise, the performance of all the baseline methods is degraded as the noise ratio increases, which is in accordance with our assumption. In addition, we also find that symmetric noise leads to larger decrease in the performance compared to asymmetric noise in both $5$-way $1$-way and $10$-way $1$-way tasks. The main reason is that corrupting a label to a wider range of node classes may lead to a more challenging weakly-supervised meta-learning task. However, when we increase the noise ratio for either the symmetric or asymmetric noise, the performance of Meta-GIN does not decrease very much. It can obtain larger improvement compared with the baselines in the data with higher noise level. This verifies the effectiveness of Meta-GIN in achieving robust performance on weakly-labeled data.

\begin{figure}[t]
\vspace{-0.05in}
    \graphicspath{{figures/}}
    \centering
    \subfigure[10-w 1-s (Sym $\epsilon = 30\%$)] 
    {
    \includegraphics[width=0.45\columnwidth]{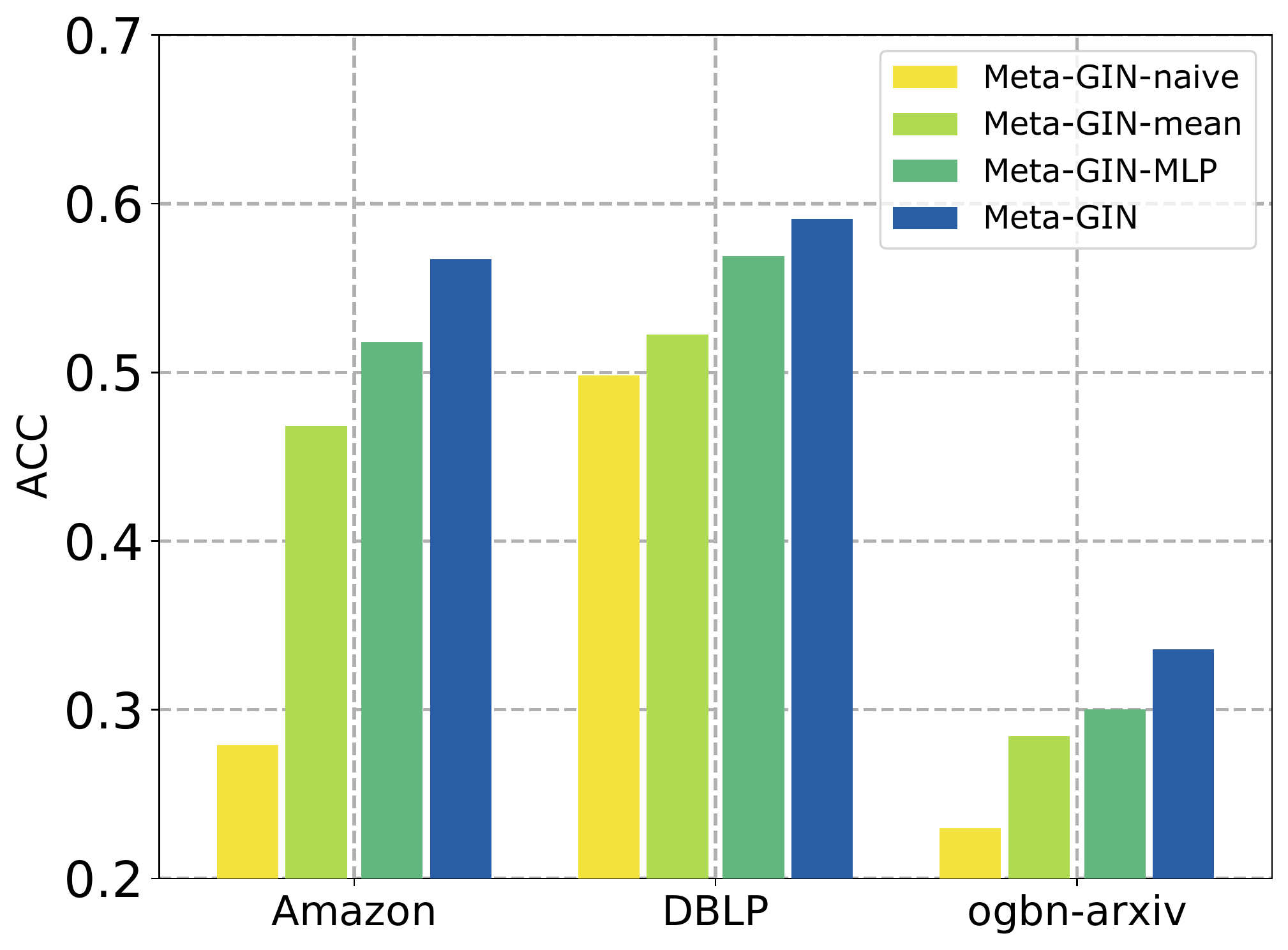}
    }
    \hspace{-0.1cm}
    \subfigure[10-w 1-s (Asym $\epsilon = 30\%$)]
    {
    \includegraphics[width=0.45\columnwidth]{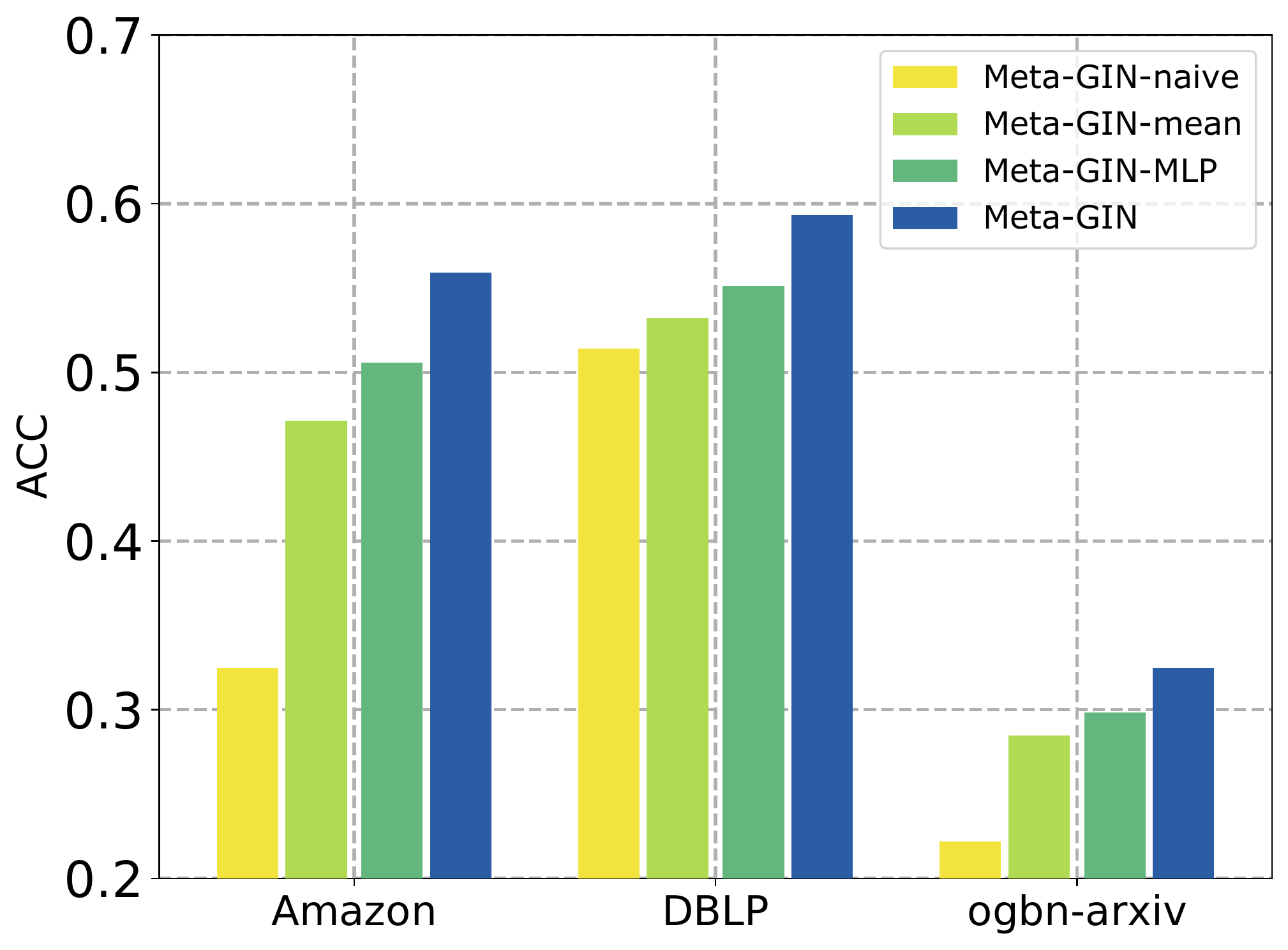}
    }
     \vspace{-0.1in}
    \caption{Ablation results for different model variants.}%

    \label{fig:ablation}
\end{figure} 

\label{sec:parameter}  
\smallskip
\noindent{\textbf{Ablation Study.}} To investigate the contribution of each component in Meta-GIN, we compare it with its variants in Figure \ref{fig:ablation}. Specifically, \textit{Meta-GIN-naive} can be considered as a naive variant by excluding both robustness-enhanced episodic training and node interpolation. \textit{Meta-GIN-MLP} and \textit{Meta-GIN-mean} denote the variants that calculate the confidence score for each node using a fully connecting layer and taking the average, respectively. As shown in the reported results, \textit{Meta-GIN-naive} is highly vulnerable to label noise and unable to obtain competitive results with other variants on weakly-labeled few-shot node classification. Based on the proposed robustness-enhanced episodic training, \textit{Meta-GIN-mean} uses the simplest way to compute the noise-reduced node representations, but can significantly outperforms \textit{Meta-GIN-naive}, which verifies the importance of using the new episodic training paradigm. Meanwhile, though \textit{Meta-GIN-MLP} can improve \textit{Meta-GIN-mean} by assigning weigthed confidence score to each node, it still fall behind our approach, which shows the node interpolation module can better estimate the confidence score of each weakly-labeled node via message passing. 

\begin{figure}
\vspace{-0.15in}
    \graphicspath{{figures/}}
    \centering
    \subfigure[$5$-w $K$-s (Sym $\epsilon = 30\%$)] 
    {
    \includegraphics[width=0.44\columnwidth]{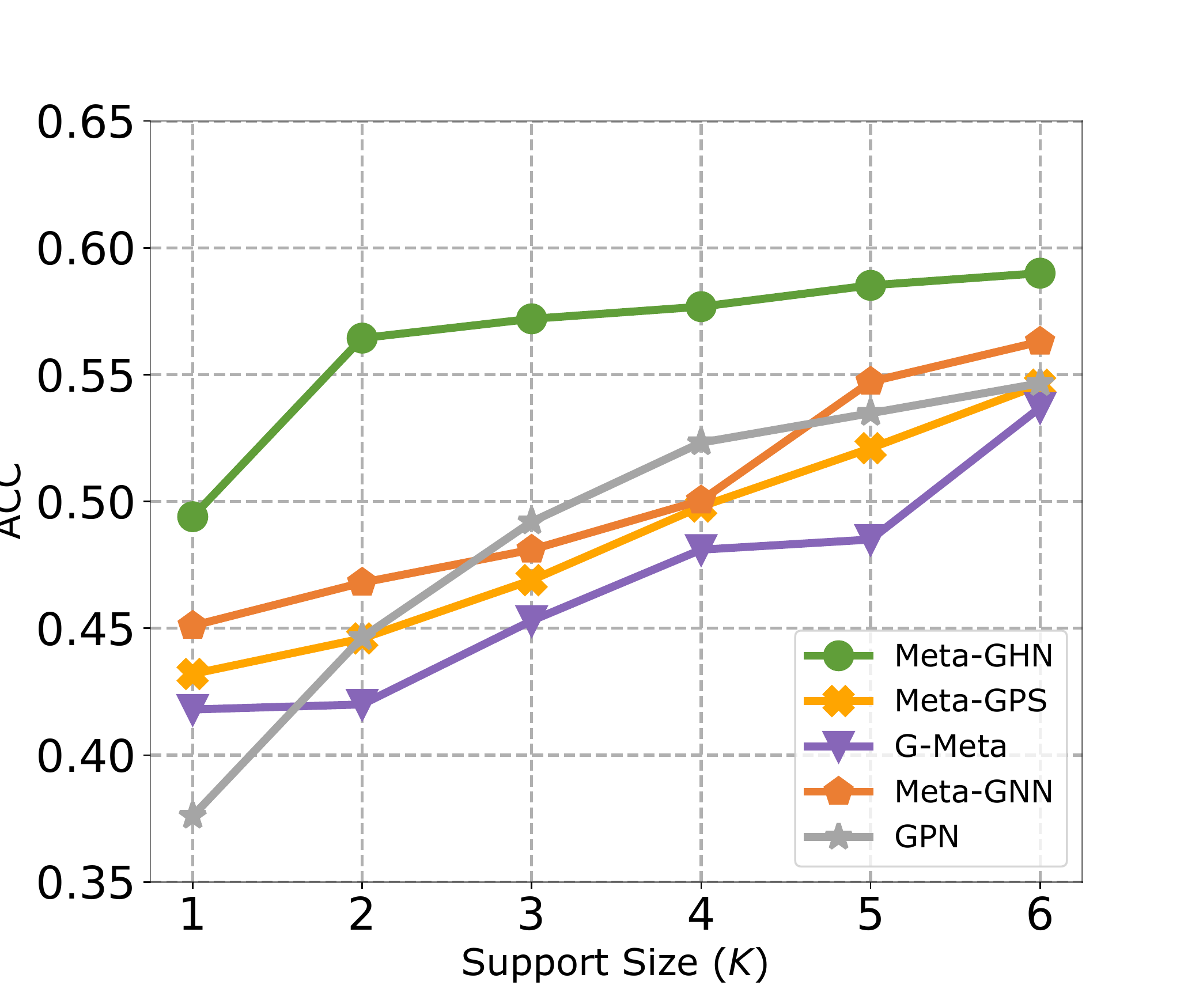}
    }
    \hspace{-0.1cm}
    \subfigure[$5$-w $1$-s (Sym $\epsilon = 30\%$)]
    {
    \includegraphics[width=0.44\columnwidth]{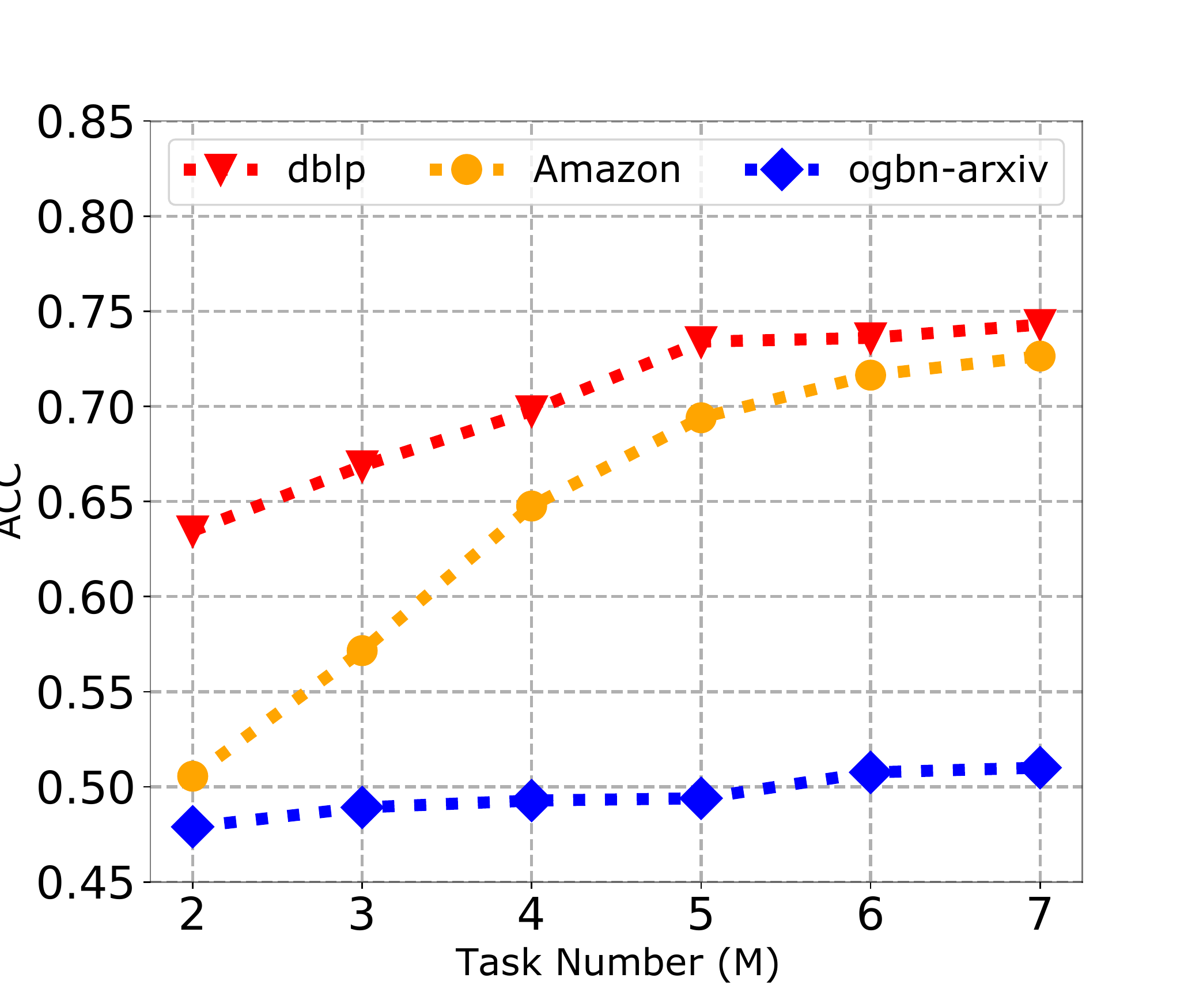}
    }
    \vspace{-0.1in}
    \caption{Evaluation for different parameters .}
    \label{fig:setsz}
\end{figure}

\smallskip
\noindent{\textbf{Parameter Analysis.}} To further understand the model design, we analyze the sensitivity of Meta-GIN to the support size $K$ and the task number $M$. Due to the space limit, here we show the results under the task of $5$-way $1$-shot with symmetric noise ($\epsilon = 0.3 $), similar patterns can be observed for other cases. In Figure \ref{fig:setsz} (a),
we summarize the performance of Meta-GIN with various support size $K$ on ogbn-arxiv and we can observe that the proposed Meta-GIN can always achieve the best performance on different $5$-way $K$-shot tasks. This demonstrates the superiority of Meta-GIN for solving weakly-supervised graph few-shot learning problems. Next, we investigate the performance of Meta-GIN by varying the task number $M$ and show the results of $5$-way $1$-shot (Sym) on the three datasets. From Figure \ref{fig:setsz} (b), we find that by increasing $M$, the performance of Meta-GIN gradually improves, which indicates that interpolating more tasks is helpful for noise-reduced node representations. Also, the model performance become stable when $M \geq 5$, thus $5$ is the appropriate value for $M$ to obtain satisfying performance considering both efficiency and effectiveness.

%% file: Conclusion.tex
In this paper, we introduce a novel graph meta-learning framework Graph Interpolation Networks (Meta-GIN) to solve few-shot learning problems under the weakly-supervised setting. Unlike existing methods, our approach does not require abundant golden labeled data from seen classes and can be meta-learned to denoise for extracting highly transferable meta-knowledge from weakly-labeled data. Essentially, Meta-GHN leverages robustness-enhanced episodic training to interpolate node representations by comparing and summarizing from weakly-labeled data in a meta-learning fashion. The empirical results over different datasets demonstrate that our proposed model can effectively generalize to unseen tasks.


%% file: Appendix.tex
\section{REPRODUCIBILITY SUPPLEMENTARY}
\subsection{Datasets Details and Preprocessing}
\label{sec:preprocess}


\smallskip
\noindent\textbf{Amazon.} To construct the graph, we retrieved all the products belonging to the top-level category ``Electronics'' from the publis Amazon dataset\footnote{\url{http://snap.stanford.edu/data/amazon/productGraph/}}. We apply bag-of-words model on product descriptions to obtain node attributes and usethe complementary relationship (``bought together'') between products to connect the nodes. Additionally, each product corresponds to a low-level category, e.g., Monopods, LED TVs and DVD Recorders, which is used to define the node label. Those classes containing 100 to 1000 nodes are selected and the isolated products is deleted.  

\noindent\textbf{DBLP.} This public citation data\footnote{\url{https://lfs.aminer.cn/misc/dblp.v11.zip}} is extracted from DBLP, containing the available papers with the corresponding abstract, authors, references and venue. In the experiment, we filter out venues which lasted for less than 20 years and focus on venues which has published 100 to 1000 papers. For constructing the graph, each paper in these venues is treated as a node and the citation relations are regarded as links between them. We apply bag-of-words model on the paper abstract to obtain node attributes. 

\noindent\textbf{ogbn-arxiv.} This citation graph is directly obtained from Open Graph Benchmark (OGB), which is a collection of benchmark datasets for graph machine learning research\footnote{\url{https://ogb.stanford.edu/}}. Specifically, ogbn-arxiv is built with all Computer Science Arxiv papers indexed by MAG. For each paper, its attributes are obtained by averaging the 128-dimension word2vec embeddings of words in its title and abstract. 


\begin{table}[!h]
\caption{Statistics of the evaluation datasets.}
\centering
\setlength{\tabcolsep}{3pt}
\scalebox{0.9}{
\begin{tabular}{@{}lcccc@{}}
\toprule

\textbf{Datasets} & \# nodes & \# edges  & \# attributes & \# Train/Valid/Test  \\ \midrule
Amazon & 42,318  & 43,556   & 8,669 & 90/37/40\\
DBLP & 40,672 & 288,270   & 7,202   & 80/27/30\\
ogbn-arxiv & 169,343  & 1,166,243  & 128 & 16/12/12\\
\bottomrule
\end{tabular}}
\label{tab:datasets}
\end{table}

\subsection{Label Noise Injection}
\label{sec:noise}
To enable our experiments on few-shot node classification with weakly-labeled data, we need to inject label noise to the training data. We focus on two representative types of label noise: symmetric and asymmetric noise \cite{chen2019understanding}. In fact, the noise injection can be done by flipping the labels following the transition probabilities defined in a corruption matrix $T$, in which $T_{ij}$ denotes the probability of flipping class $c_i$ to class $c_j$. In Figure \ref{fig:corruption_matrix}, we visualize the examples of corruption matrix for the dataset containing 5 label classes. For injecting noise of ratio $\epsilon$ to a dataset with $P$ classes: 
\begin{itemize}[leftmargin=*]
    \item \textbf{Symmetric noise} flips a label uniformly to all the other classes, s.t. $T_{ii} = 1-\epsilon$ and $T_{ij} = \epsilon/(P-1)$ if $i \neq j$. 

    \item  \textbf{Asymmetric noise} flips a label to a different class with probability $\epsilon$, s.t. $T_{ii} = 1-\epsilon$ and $\exists i \neq j$, $T_{ij} = \epsilon$. 
\end{itemize}

\subsection{Implementation of Baselines}
\label{sec:reproducibility}


%

We randomly sampled 100 \textit{meta-test} tasks from the test node classes and evaluate both Meta-GIN and the baseline methods on these tasks. The process is repeated for 10 times to obtain the reported results in the paper.
We test all the baseline methods with the publicly released implementations. In the experiments, we fine-tune the hyperparameters to report their best performance.
\begin{itemize}[leftmargin=*]
    \item \textbf{GCN.} It utilizes two graph convolutional layers (32, N dimensions) to learn the node representations for $N$-way-$K$-shot tasks.
    
    \item  \textbf{SGC.} This linear model reduces the unnecessary complexity of GCN by successively collapsing the convolution functions between consecutive layers into a linear transformation. After the feature pre-processing step, it learns the node representations with 2-layer feature propagation. 
    
    \item  \textbf{GraphSAGE.} It can efficiently generate node embeddings by uniformly sampling a fixed size of neighbors and then aggregating the feature information from the neighbors. We set the search depth to 2 and the neighborhood sample size to 35 for both layers. ReLu function is used for non-linearity and the mean-pooling aggregator is selected for comparison. Two layers (32, N dimensions) are used for learning node representations.
    
    \item \textbf{PTA.} It is a decoupled GNN which is robust to label noise using adaptive weighting strategy. As suggested in the original paper, we use $K = 10$ propagation steps and set the optimal teleport probability $\alpha = 0.1$. For fair comparison, it employs a two-layer (32, N dimensions) MLP for representation learning.

    \item \textbf{Meta-GNN.} It applies MAML~\cite{finn2017model} to Simple Graph Convolution (SGC) for few-shot node classification in graphs. A 2-layer SGC is used for network embedding. Each batch contains 5 tasks. As suggested in the original paper, we set task-learning rate $\alpha_1 = 0.5$ and meta-learning rate $\alpha_2 = 0.003$. 
    
    \item \textbf{GPN.} This Graph Prototypical Network can learn highly representative class prototypes with a GNN-based network encoder and node valuator. It predicts node labels by measuring their similarity with prototypes. It employs a 2-layer (32, 16 dimensions) GCN as network encoder. The node valuator relies on two score aggregation layers. We use the optimal learning rate $\alpha = 0.005$.  
    
    \item \textbf{G-Meta.}  It constructs a local subgraph for each node and regards the centroid embedding of subgraphs as the prototypes. It is optimized with both the prototypical loss and MAML. It takes the 2-hop neighbors into consideration and employs two graph convolutional aggregation layers (32, 32 dimensions) for representation learning. We select the optimal update learning rate $\alpha = 0.01$ meta learning rate $\beta = 0.001$. 

    \item \textbf{Meta-GPS.} This graph Meta-learning framework is based on Prototype and Scaling \& shifting transformation to obtain highly transferable meta-knowledge from meta-training tasks. It employs a 2-layer (32, 16 dimensions) network encoder. We use the optimal step size $\alpha = 0.5$, meta-learning rate $\beta =0.001$ and regularization coefficient $\gamma = 0.001$.

    \begin{figure}
    \graphicspath{{figures/}}
    \centering
    \subfigure[\textbf{Sym Noise ($\epsilon = 0.3$)}] 
    {
    \includegraphics[width=0.485\columnwidth]{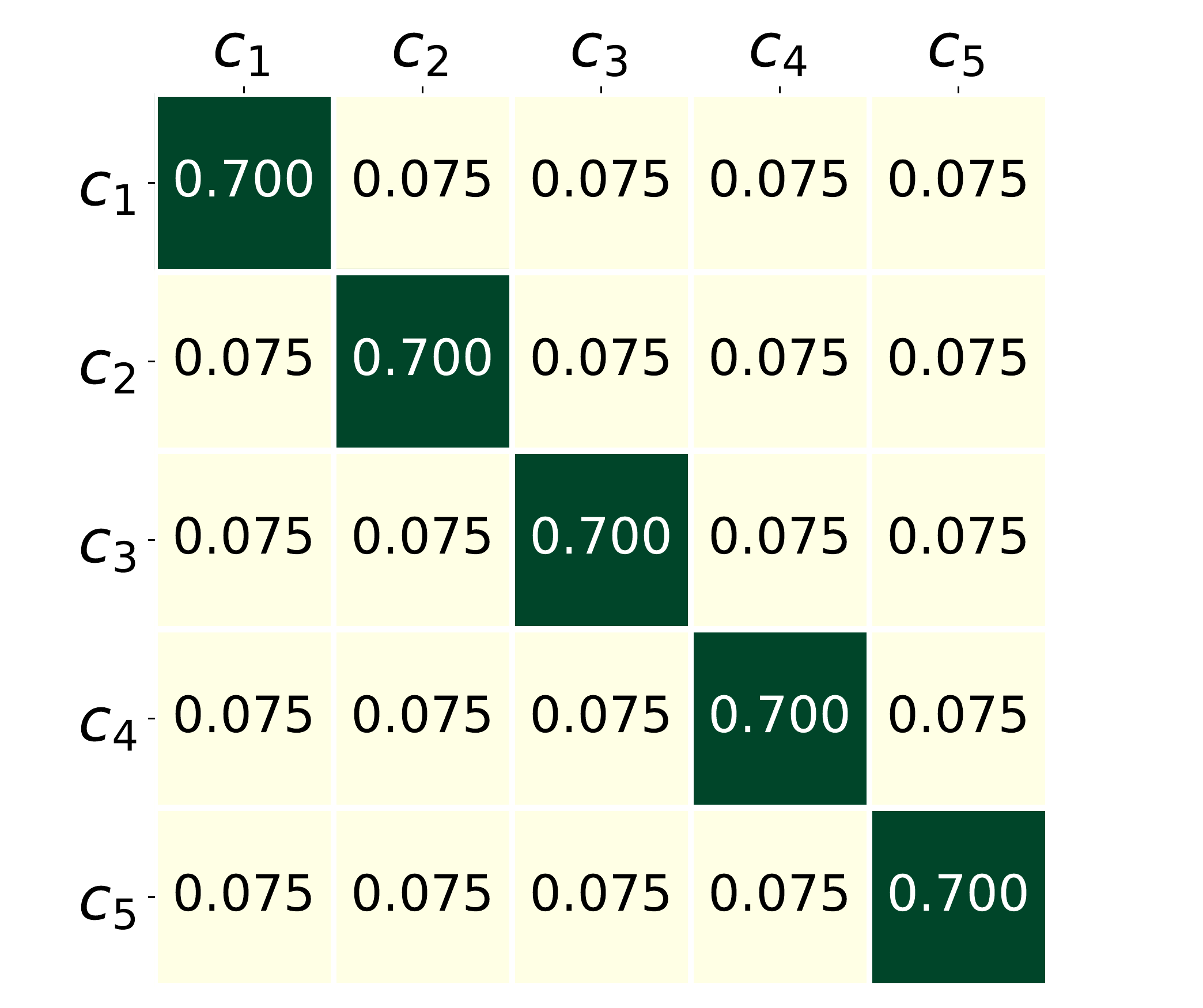}
    }
    \hspace{-0.5cm}
    \subfigure[\textbf{Asym Noise ($\epsilon = 0.3$)}]
    {
    \includegraphics[width=0.485\columnwidth]{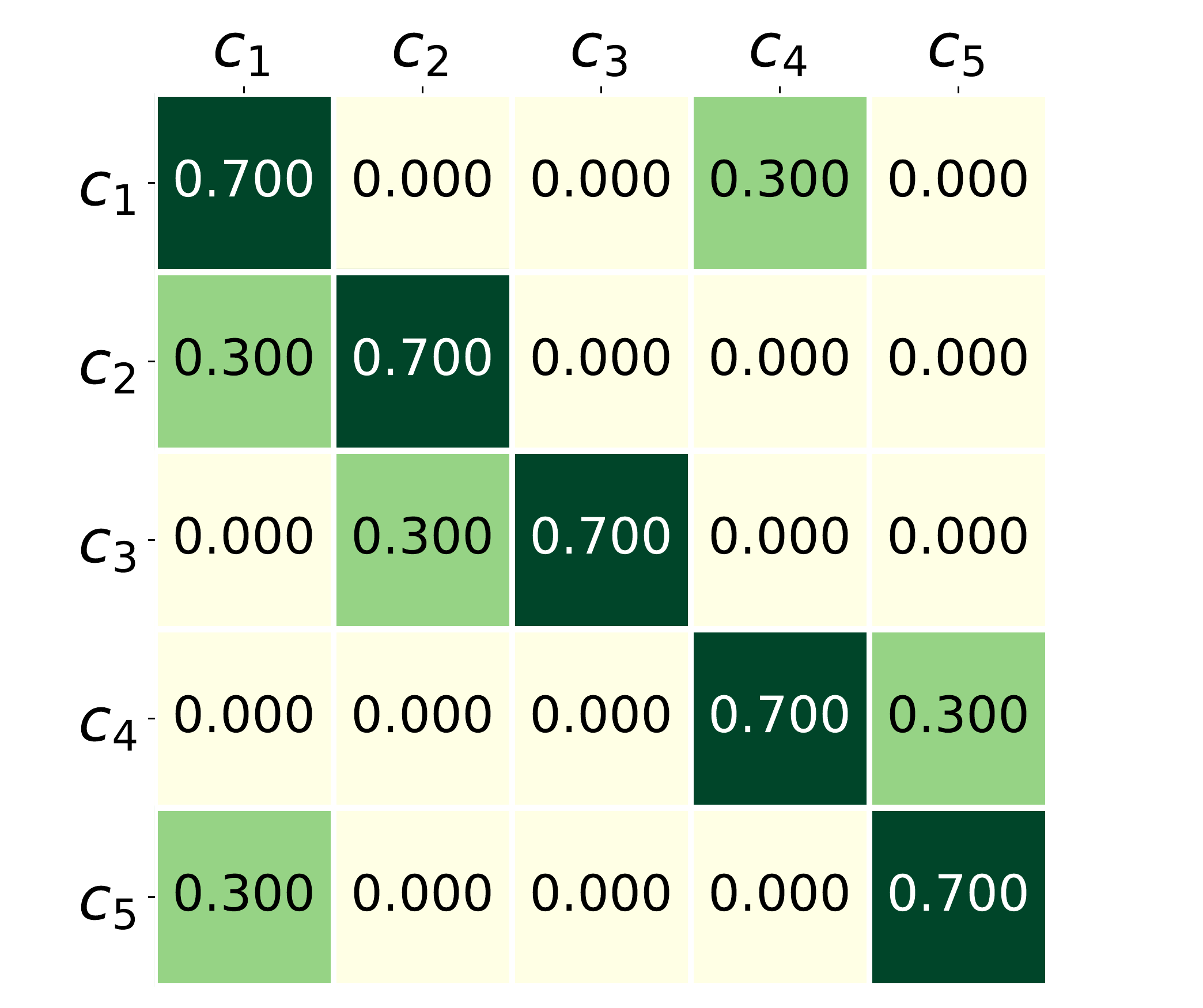}
    }
    \caption{Example of the noise corruption matrix.}%

    \label{fig:corruption_matrix}
\end{figure}

\end{itemize}

\smallskip
\noindent\textbf{Model Implementation.} The proposed model is implemented in PyTorch. Specifically, we employ a 2-layer propagation SGC for the node representation learning module. As for the node interpolation module, we use one aggregation layer and the negative slope for the LeakyReLU in it is set to be $0.2$. We grid search for task numbers in \{1, 5, 10, 15, 20, 25\}, meta learning rate $\alpha$ and meta step size $\beta$ in \{$0.0001$, $0.0005$, $0.001$, $0.005$, $0.01$, $0.05$, $0.1$, $0.5$\}. The optimal values are selected when the model achieve the best performance for validation set. We select the meta-learning rate $\alpha$ to be $0.1$ and the meta step size $\beta$ to be $0.001$. For model training, the task number in each batch is $5$ and the query size $K'$ is $5$. For constructing the meta-training episodes, unless otherwise notice, we let the set size $M$ to be $5$ for both the support and query set. We train the model with 20,000 episodes or stop earlier when the performance on validation set converges. In the testing phase, we feed the $N$-way $K$-shot support set from unseen classes to both Meta-GIN and the baseline models for fair comparison. 

\smallskip
\noindent\textbf{Packages Used for Implementation.} For reproducibility, we also list the packages we use in the implementation with their corresponding versions: 
 \begin{itemize}
     \item python==3.7.9
     \item pytorch==1.4.0
     \item cuda==10.1    
     \item numpy==1.19.2
     \item scipy==1.6.0
     \item scikit-learn==0.24.0
 \end{itemize}

%% file: Main.bbl
\begin{thebibliography}{32}
\providecommand{\natexlab}[1]{#1}

\bibitem[{Baek, Lee, and Hwang(2020)}]{baek2020learning}
Baek, J.; Lee, D.~B.; and Hwang, S.~J. 2020.
\newblock Learning to Extrapolate Knowledge: Transductive Few-shot Out-of-Graph
  Link Prediction.
\newblock \emph{NeurIPS}.

\bibitem[{Cao, Lu, and Xu(2016)}]{cao2016deep}
Cao, S.; Lu, W.; and Xu, Q. 2016.
\newblock Deep neural networks for learning graph representations.
\newblock In \emph{AAAI}.

\bibitem[{Chang et~al.(2015)Chang, Han, Tang, Qi, Aggarwal, and
  Huang}]{chang2015heterogeneous}
Chang, S.; Han, W.; Tang, J.; Qi, G.-J.; Aggarwal, C.~C.; and Huang, T.~S.
  2015.
\newblock Heterogeneous network embedding via deep architectures.
\newblock In \emph{KDD}.

\bibitem[{Chauhan, Nathani, and Kaul(2019)}]{chauhan2019few}
Chauhan, J.; Nathani, D.; and Kaul, M. 2019.
\newblock FEW-SHOT LEARNING ON GRAPHS VIA SUPER-CLASSES BASED ON GRAPH SPECTRAL
  MEASURES.
\newblock In \emph{ICLR}.

\bibitem[{Chen et~al.(2019)Chen, Liao, Chen, and Zhang}]{chen2019understanding}
Chen, P.; Liao, B.~B.; Chen, G.; and Zhang, S. 2019.
\newblock Understanding and utilizing deep neural networks trained with noisy
  labels.
\newblock In \emph{ICML}.

\bibitem[{Ding et~al.(2019)Ding, Li, Bhanushali, and Liu}]{ding2019deep}
Ding, K.; Li, J.; Bhanushali, R.; and Liu, H. 2019.
\newblock Deep anomaly detection on attributed networks.
\newblock In \emph{SDM}.

\bibitem[{Ding et~al.(2020)Ding, Wang, Li, Shu, Liu, and Liu}]{ding2020graph}
Ding, K.; Wang, J.; Li, J.; Shu, K.; Liu, C.; and Liu, H. 2020.
\newblock Graph prototypical networks for few-shot learning on attributed
  networks.
\newblock In \emph{CIKM}.

\bibitem[{Dong et~al.(2021)Dong, Chen, Feng, He, Bi, Ding, and
  Cui}]{dong2021equivalence}
Dong, H.; Chen, J.; Feng, F.; He, X.; Bi, S.; Ding, Z.; and Cui, P. 2021.
\newblock On the Equivalence of Decoupled Graph Convolution Network and Label
  Propagation.
\newblock In \emph{The Web Conference}.

\bibitem[{Finn, Abbeel, and Levine(2017)}]{finn2017model}
Finn, C.; Abbeel, P.; and Levine, S. 2017.
\newblock Model-agnostic meta-learning for fast adaptation of deep networks.
\newblock In \emph{ICML}.

\bibitem[{Hamilton, Ying, and Leskovec(2017)}]{hamilton2017inductive}
Hamilton, W.; Ying, Z.; and Leskovec, J. 2017.
\newblock Inductive representation learning on large graphs.
\newblock In \emph{NeurIPS}.

\bibitem[{Hendrycks et~al.(2018)Hendrycks, Mazeika, Wilson, and
  Gimpel}]{hendrycks2018using}
Hendrycks, D.; Mazeika, M.; Wilson, D.; and Gimpel, K. 2018.
\newblock Using trusted data to train deep networks on labels corrupted by
  severe noise.
\newblock In \emph{NeuIPS}.

\bibitem[{Huang and Zitnik(2020)}]{huang2020graph}
Huang, K.; and Zitnik, M. 2020.
\newblock Graph meta learning via local subgraphs.
\newblock In \emph{NeurIPS}.

\bibitem[{Jiang et~al.(2018)Jiang, Zhou, Leung, Li, and
  Fei-Fei}]{jiang2018mentornet}
Jiang, L.; Zhou, Z.; Leung, T.; Li, L.-J.; and Fei-Fei, L. 2018.
\newblock Mentornet: Learning data-driven curriculum for very deep neural
  networks on corrupted labels.
\newblock In \emph{ICML}.

\bibitem[{Kipf and Welling(2017)}]{kipf2017semi}
Kipf, T.~N.; and Welling, M. 2017.
\newblock Semi-supervised classification with graph convolutional networks.
\newblock In \emph{NeurIPS}.

\bibitem[{Lan et~al.(2020)Lan, Wang, Du, Song, Tao, and Guan}]{lan2020node}
Lan, L.; Wang, P.; Du, X.; Song, K.; Tao, J.; and Guan, X. 2020.
\newblock Node Classification on Graphs with Few-Shot Novel Labels via Meta
  Transformed Network Embedding.
\newblock In \emph{NeurIPS}.

\bibitem[{Liu et~al.(2022)Liu, Li, Li, Giunchiglia, Feng, and
  Guan}]{liu2022few}
Liu, Y.; Li, M.; Li, X.; Giunchiglia, F.; Feng, X.; and Guan, R. 2022.
\newblock Few-shot Node Classification on Attributed Networks with Graph
  Meta-learning.
\newblock In \emph{SIGIR}.

\bibitem[{Liu et~al.(2021)Liu, Fang, Liu, and Hoi}]{liu2021relative}
Liu, Z.; Fang, Y.; Liu, C.; and Hoi, S.~C. 2021.
\newblock Relative and Absolute Location Embedding for Few-Shot Node
  Classification on Graph.
\newblock In \emph{AAAI}.

\bibitem[{Liu et~al.(2020)Liu, Zhang, Fang, Zhang, and Hoi}]{liu2020towards}
Liu, Z.; Zhang, W.; Fang, Y.; Zhang, X.; and Hoi, S.~C. 2020.
\newblock Towards locality-aware meta-learning of tail node embeddings on
  networks.
\newblock In \emph{CIKM}.

\bibitem[{Ma et~al.(2020)Ma, Bu, Yang, Zhang, Yao, Yu, Zhou, and
  Yan}]{ma2020adaptive}
Ma, N.; Bu, J.; Yang, J.; Zhang, Z.; Yao, C.; Yu, Z.; Zhou, S.; and Yan, X.
  2020.
\newblock Adaptive-Step Graph Meta-Learner for Few-Shot Graph Classification.
\newblock In \emph{CIKM}.

\bibitem[{McAuley, Pandey, and Leskovec(2015)}]{mcauley2015inferring}
McAuley, J.; Pandey, R.; and Leskovec, J. 2015.
\newblock Inferring networks of substitutable and complementary products.
\newblock In \emph{KDD}.

\bibitem[{Ren et~al.(2018{\natexlab{a}})Ren, Triantafillou, Ravi, Snell,
  Swersky, Tenenbaum, Larochelle, and Zemel}]{ren2018meta}
Ren, M.; Triantafillou, E.; Ravi, S.; Snell, J.; Swersky, K.; Tenenbaum, J.~B.;
  Larochelle, H.; and Zemel, R.~S. 2018{\natexlab{a}}.
\newblock Meta-learning for semi-supervised few-shot classification.
\newblock In \emph{ICLR}.

\bibitem[{Ren et~al.(2018{\natexlab{b}})Ren, Zeng, Yang, and
  Urtasun}]{ren2018learning}
Ren, M.; Zeng, W.; Yang, B.; and Urtasun, R. 2018{\natexlab{b}}.
\newblock Learning to reweight examples for robust deep learning.
\newblock In \emph{ICML}.

\bibitem[{Snell, Swersky, and Zemel(2017)}]{snell2017prototypical}
Snell, J.; Swersky, K.; and Zemel, R. 2017.
\newblock Prototypical networks for few-shot learning.
\newblock In \emph{NeurIPS}.

\bibitem[{Tang et~al.(2008)Tang, Zhang, Yao, Li, Zhang, and
  Su}]{tang2008arnetminer}
Tang, J.; Zhang, J.; Yao, L.; Li, J.; Zhang, L.; and Su, Z. 2008.
\newblock Arnetminer: extraction and mining of academic social networks.
\newblock In \emph{KDD}.

\bibitem[{Veli{\v{c}}kovi{\'c} et~al.(2018)Veli{\v{c}}kovi{\'c}, Cucurull,
  Casanova, Romero, Lio, and Bengio}]{velickovic2017graph}
Veli{\v{c}}kovi{\'c}, P.; Cucurull, G.; Casanova, A.; Romero, A.; Lio, P.; and
  Bengio, Y. 2018.
\newblock Graph attention networks.
\newblock In \emph{ICLR}.

\bibitem[{Vinyals et~al.(2016)Vinyals, Blundell, Lillicrap, Wierstra
  et~al.}]{vinyals2016matching}
Vinyals, O.; Blundell, C.; Lillicrap, T.; Wierstra, D.; et~al. 2016.
\newblock Matching networks for one shot learning.
\newblock In \emph{NeurIPS}.

\bibitem[{Wu et~al.(2019)Wu, Zhang, Souza~Jr, Fifty, Yu, and
  Weinberger}]{wu2019simplifying}
Wu, F.; Zhang, T.; Souza~Jr, A. H.~d.; Fifty, C.; Yu, T.; and Weinberger, K.~Q.
  2019.
\newblock Simplifying graph convolutional networks.
\newblock \emph{ICML}.

\bibitem[{Yao et~al.(2020)Yao, Zhang, Wei, Jiang, Wang, Huang, Chawla, and
  Li}]{yao2019graph}
Yao, H.; Zhang, C.; Wei, Y.; Jiang, M.; Wang, S.; Huang, J.; Chawla, N.~V.; and
  Li, Z. 2020.
\newblock Graph few-shot learning via knowledge transfer.
\newblock In \emph{AAAI}.

\bibitem[{Zhang et~al.(2020)Zhang, Yao, Huang, Jiang, Li, and
  Chawla}]{zhang2020few}
Zhang, C.; Yao, H.; Huang, C.; Jiang, M.; Li, Z.; and Chawla, N.~V. 2020.
\newblock Few-shot knowledge graph completion.
\newblock In \emph{AAAI}.

\bibitem[{Zhang et~al.(2019)Zhang, Zhao, Ni, Xu, and
  Yang}]{zhang2019variational}
Zhang, J.; Zhao, C.; Ni, B.; Xu, M.; and Yang, X. 2019.
\newblock Variational few-shot learning.
\newblock In \emph{ICCV}.

\bibitem[{Zhang, Wang, and Qiao(2019)}]{zhang2019metacleaner}
Zhang, W.; Wang, Y.; and Qiao, Y. 2019.
\newblock Metacleaner: Learning to hallucinate clean representations for
  noisy-labeled visual recognition.
\newblock In \emph{CVPR}.

\bibitem[{Zhou et~al.(2019)Zhou, Cao, Zhang, Trajcevski, Zhong, and
  Geng}]{zhou2019meta}
Zhou, F.; Cao, C.; Zhang, K.; Trajcevski, G.; Zhong, T.; and Geng, J. 2019.
\newblock Meta-GNN: On Few-shot Node Classification in Graph Meta-learning.
\newblock In \emph{CIKM}.

\end{thebibliography}
